\documentclass[10pt]{article} 

\usepackage[accepted]{rlj} 

\usepackage{amssymb}            
\usepackage{mathtools}          
\usepackage{mathrsfs}           
\usepackage{graphicx}           
\usepackage{subcaption}         
\usepackage[space]{grffile}     
\usepackage{url}                
\usepackage{lipsum}             
\usepackage{wrapfig}
\usepackage{listings}
\usepackage[frozencache=true,cachedir=.]{minted}
\usepackage{natbib}
\usepackage{subcaption}
\usepackage{booktabs}

\usepackage{etoolbox}
\makeatletter
\patchcmd{\hyper@makecurrent}{%
    \ifx\Hy@param\Hy@chapterstring
        \let\Hy@param\Hy@chapapp
    \fi
}{%
    \iftoggle{inappendix}{
        \@checkappendixparam{chapter}%
        \@checkappendixparam{section}%
        \@checkappendixparam{subsection}%
        \@checkappendixparam{subsubsection}%
        \@checkappendixparam{paragraph}%
        \@checkappendixparam{subparagraph}%
    }{}%
}{}{\errmessage{failed to patch}}

\newcommand*{\@checkappendixparam}[1]{%
    \def\@checkappendixparamtmp{#1}%
    \ifx\Hy@param\@checkappendixparamtmp
        \let\Hy@param\Hy@appendixstring
    \fi
}
\makeatletter

\newtoggle{inappendix}
\togglefalse{inappendix}

\apptocmd{\appendix}{\toggletrue{inappendix}}{}{\errmessage{failed to patch}}

\title{Syllabus: Portable Curricula for Reinforcement Learning Agents}

\setrunningtitle{Syllabus: Portable Curricula for Reinforcement Learning Agents}

\author{Ryan Sullivan \textsuperscript{1},  Ryan Pégoud \textsuperscript{2}, Ameen Ur Rehman \textsuperscript{3}, Xinchen Yang \textsuperscript{1}, Junyun Huang \textsuperscript{1}, Aayush Verma \textsuperscript{1}, Nistha Mitra \textsuperscript{1}, John P. Dickerson \textsuperscript{1}}

\coverPageAuthor{Ryan Sullivan, Ryan Pégoud, Ameen Ur Rehman, Xinchen Yang, Junyun Huang, Aayush Verma, Nistha Mitra, John P. Dickerson}

\emails{rsulli@umd.edu}

\affiliations{
$^{1}$\textbf{University of Maryland, College Park} \hspace{0.5em}
$^{2}$\textbf{University College London} \\
$^{3}$\textbf{Jamia Hamdard University}
}

\contribution{
    This paper introduces Syllabus, a library of portable curriculum learning algorithms and infrastructure for synchronizing curricula across reinforcement learning environments running in separate processes. Syllabus includes portable implementations of several popular automatic curriculum learning algorithms and tools for manually designing curricula.
    }
    {
    There are open-source curriculum learning libraries \citep{dcd, jiang2023minimax, dharna2022watts, coward2024jaxued}, but they build curriculum logic into the RL training code, making it difficult to extend methods and apply them to new environments. Syllabus is the first portable infrastructure for curriculum learning.
    }

\contribution{
    We evaluate tuned curriculum learning baselines in 4 environments including 2 which have not previously been explored in the context of curriculum learning. These baselines provide a solid foundation for future curriculum learning research.
    }
    {
    We implement baselines from \citet{jiang2021prioritized}, \citet{kanitscheider2021multi}, \citet{zhang2023omni}, and \citet{rutherfordno} and apply all four algorithms to baseline environments used in these works. Some of these evaluations are reproductions of the experiments in those papers, but most are novel.
    }

\contribution{
      Our experiments demonstrate that popular curriculum learning methods are far less effective outside of the environments in which they were originally developed, and that more advanced methods may be necessary in complex environments.
    }
    {
    Previous work successfully used automatic curricula over level seeds to train agents in procedurally generated environments. We show that these curricula over level seeds are ineffective in new or complex environments.
    }


\summary{Curriculum learning has been a quiet, yet crucial component of many high-profile successes of reinforcement learning. Despite this, it is still a niche topic that is not directly supported by any of the major reinforcement learning libraries. These methods can improve the capabilities and generalization of RL agents, but often require complex changes to training code, limiting their impact on the field. We introduce Syllabus, a portable curriculum learning library, as a solution to this problem. Syllabus provides a universal API for curriculum learning, modular implementations of popular automatic curriculum learning methods, and infrastructure that allows them to be easily integrated with asynchronous training code in nearly any RL library. Syllabus provides a minimal API for core curriculum learning components, making it easier to design new algorithms and adapt existing ones to new environments. We demonstrate this by evaluating the algorithms in Syllabus on several new environments, each using agents written in a different RL library. We present the first examples of automatic curriculum learning in NetHack and Neural MMO, two of the most challenging RL benchmarks, and find evidence that existing methods do not easily transfer to complex new environments. Syllabus can be found at \href{https://github.com/RyanNavillus/Syllabus}{https://github.com/RyanNavillus/Syllabus}.
}

\begin{document}

\makeCover
\maketitle

\begin{abstract}
    Curriculum learning has been a quiet, yet crucial component of many high-profile successes of reinforcement learning. Despite this, it is still a niche topic that is not directly supported by any of the major reinforcement learning libraries. These methods can improve the capabilities and generalization of RL agents, but often require complex changes to training code. We introduce Syllabus, a portable curriculum learning library, as a solution to this problem. Syllabus provides a universal API for curriculum learning, modular implementations of popular automatic curriculum learning methods, and infrastructure that allows them to be easily integrated with asynchronous training code in nearly any RL library. Syllabus provides a minimal API for core curriculum learning components, making it easier to design new algorithms and adapt existing ones to new environments. We demonstrate this by evaluating the algorithms in Syllabus on several new environments, each using agents written in a different RL library. We present the first examples of automatic curriculum learning in NetHack and Neural MMO, two of the most challenging RL benchmarks, and find evidence that existing methods do not directly transfer to complex new environments. Syllabus can be found at \href{https://github.com/RyanNavillus/Syllabus}{https://github.com/RyanNavillus/Syllabus}.
\end{abstract}

\section{Introduction}
\label{sec:introduction}

Curricula have been a core component of many of the successes of reinforcement learning (RL). AlphaGo \citep{alphago} was trained with self-play, AlphaStar used a novel league training method to achieve grandmaster level play in Starcraft II \citep{Vinyals2019GrandmasterLI}, and GT Sophy \citep{gtsophy} was taught to outrace professionals in Gran Turismo with manually curated sections of race tracks. Curriculum learning is essential in environments with large task spaces where many tasks are too challenging or too easy to provide a useful learning signal. In these settings agents must prioritize tasks that teach transferable skills to accelerate learning in new tasks. This problem is especially pronounced in open-ended environments with infinite or evolving task spaces, similar to the real world. Open-endedness research seeks to co-evolve agents and environments in order to create more complex tasks and incentivize more intelligent agent behavior \citep{wang2019paired, wang2020enhanced}. To approximate this complexity, the field focuses on the most challenging RL benchmarks including Minecraft \citep{guss2019minerl, fan2022minedojo}, NetHack \citep{kuttler2020nethack}, and Neural MMO \citep{suarez2019neural, rosseau2022toward, suarez2024neural}. Curriculum learning is an integral component of open-ended processes, so they directly benefit from curriculum learning research.

Curriculum learning (CL) methods fit naturally into the RL framework by modifying the data distribution that an agent experiences. According to \citet{narvekar2020curriculum}, curriculum learning explores how tasks or data samples can be sequenced to learn problems that can not be solved directly. Automatic curriculum Learning (ACL) methods \citep{graves2017automated, portelas2020automatic} focus on autonomously sequencing these tasks to maximize agent performance, and have been shown to outperform random task ordering in robotics \citep{tobin2017domain, akkaya2019solving, mehta2020active}, videogames \citep{salimans2018learning, berner2019dota}, and navigation \citep{florensa2018automatic, racaniere2019automated, portelas2020teacher}. Despite the near ubiquity of curricula in successful applications of reinforcement learning, there is little support for these methods in standard RL infrastructure. In theory, CL provides orthogonal benefits to policy optimization algorithms and they can be integrated with minimal restrictions. In practice however, CL methods are diverse, complex, and necessitate significant changes to training code. CL algorithms can modify the environment configuration \citep{dennis2020emergent, jiang2021prioritized, jiang2021replay}, introduce new neural networks to train \citep{10.5555/3305890.3305962, 10.5555/3305890.3305968, 10.5555/3157096.3157262, henaff2022exploration}, or even query tasks from a large language models \citep{colas2023augmenting, du2023guiding, zhang2023omni, faldor2024omni}. Each method is different, so engineering time often scales linearly with the number of methods being evaluated.

Most open-source implementations of CL algorithms integrate their code directly within the RL algorithm, intertwining CL and RL logic. We argue that this is typically done to accommodate existing multiprocessing implementations. ACL methods maintain a sampling distribution over tasks which they update using feedback from the environment. In an asynchronous RL setting, this requires additional data exchanges between environments running on separate processes and the curriculum in the main process. The simplest solution is to add these extra messages to the existing message-passing infrastructure, usually in the \texttt{infos} dictionary. However, accessing this data requires direct changes to the training code that vary depending on the CL method, so CL algorithms are implemented as standalone libraries with custom training code. This entanglement of CL and RL code makes it difficult to isolate algorithmic details and apply CL methods beyond their original codebases, ultimately limiting reproducibility and hindering future research.

Syllabus addresses this problem by introducing a simple, portable approach to designing curriculum learning algorithms. We implement CL algorithms as modular additions to RL code, complementing their theoretical orthogonality to policy optimization. Syllabus makes minimal assumptions about the RL training code and establishes a separate synchronization pathway between the curriculum and asynchronous environments. This architecture best enables future research by integrating with existing RL infrastructure rather than attempting to replace it. We demonstrate the generality of Syllabus by implementing multiple ACL algorithms and evaluate them in several RL libraries including CleanRL \citep{huang2022cleanrl}, RLLib \citep{liang2018rllib}, moolib \citep{moolib2022}, and PufferLib \citep{Suarez23:PufferLib}. This allows us to present new baselines of several ACL algorithms on Procgen \citep{pmlr-v119-cobbe20a}, and Crafter \citep{hafner2021benchmarking}, and the first ACL results on NetHack \citep{kuttler2020nethack}, and Neural MMO \citep{suarez2019neural, suarez2024neural}.

\section{Background}
\label{sec:background}

Curriculum Learning has been studied in the context of deep supervised learning for many years \citep{10.1145/1553374.1553380, ELMAN199371}. More recently, it has been used to improve the capabilities and generalization of deep reinforcement learning agents. Curriculum learning encompasses a wide range of methods that change the training data distribution. The goal is to increase the asymptotic performance or sample efficiency of RL agents on a single environment or range of tasks by sampling tasks that provide maximal learning value. These methods often make a distinction between the goal generating teacher and the student agent that plays the tasks assigned by the teacher. \citet{narvekar2020curriculum} and \citet{portelas2020automatic} present more thorough taxonomies and surveys of existing curriculum learning methods. It can also be viewed as an extension of transfer learning, which \citet{10.5555/1577069.1755839} and \citet{zhu2023transfer} summarize in the context of RL.

Many diverse methods fall under the broad definition of curriculum learning. Some take inspiration from the Zone of Proximal Development \citep{ef4d7fb0-848f-3480-8634-d49a5f5c57df, Chaiklin_2003} which suggests that tasks in the proximal zone -- tasks that are neither too hard nor too easy -- maximize learning progress (LP). Many curriculum learning papers therefore focus on developing measures of learning progress. \citet{portelas2020teacher} fit Gaussian Mixture Models \citep{NIPS1999_97d98119} to a dataset of continuous tasks and their corresponding learning progress measures, then treat the individual Gaussians as arms in a bandit problem, allowing a teacher to bias training toward high-LP tasks. \citet{kanitscheider2021multi, tzannetos2023proximal, rutherfordno} evaluate the agent's current progress on the full task space throughout training, and compute a learning progress metric from these success rates. \citet{pmlr-v162-klink22a} introduce an optimal transport based curriculum and identify a link between interpolation based curricula and methods that use success rates, regret, or learning progress. Intrinsically motivated exploration bonuses like curiosity \citep{pathak2017curiosity} or novelty \citep{bellemare2016unifying, taiga2021bonus, henaff2022exploration} modify the reward function to induce a curriculum by incentivizing the agent to explore unseen section of the state space.

Self-play based methods \citep{5392560, 10.1145/203330.203343, silver2017mastering} create an implicit curriculum in competitive multiplayer games by training a policy against itself \citep{leibo2019autocurricula}. As the agent becomes more capable, so does the opponent, allowing it to continually improve. Training only against the current policy can lead to strategic cycles, but mixing in past policies helps prevent this \citep{brown:fp1951, Heinrich2015FictitiousSI, heinrich2016deep, lanctot2017unified, Vinyals2019GrandmasterLI}. Self-play has been extended to include multiple students \citep{sukhbaatar2018intrinsic, openai2021asymmetric} and to train a teacher represented by a neural network \citep{dutakes, dennis2020emergent, mediratta2023stabilizing}. \citet{zhang2024survey} provide an extensive survey of self-play methods in RL.

Unsupervised Environment Design (UED) is another paradigm for curriculum learning proposed by \citet{dennis2020emergent}. They differentiate UED as a framework in which environments have unspecified configuration parameters, thereby forming an Underspecified Partially Observable Markov Decision Process (UPOMDP). The parameters generate a distribution of solvable tasks, and the goal of a UED method is to train a policy that generalizes across all possible instantiations of those variables. This approach has led to several algorithms for training agents in procedurally generated games \citep{dennis2020emergent, jiang2021prioritized, jiang2021replay, beukman2024refining, pmlr-v232-mediratta23a}.

The specific policy optimization algorithm used to train agents is largely independent from the choice of curriculum. For simplicity, most curriculum learning research and all of the experiments in this paper use Proximal Policy Optimization \citep{schulman2017proximal}. PPO is an on-policy, policy gradient algorithm that has been successfully applied to a wide range of challenging RL environments. It uses a clipped objective to avoid large changes to the policy, which prevents collapses in performance and stabilizes learning \citep{pmlr-v37-schulman15, schulman2017proximal}.

\section{Related Work}
Popular RL libraries do not include curriculum learning algorithms, but there are specialized libraries for curriculum learning. The Dual Curriculum Design library \citep{jiang2021replay} incorporates multiple UED methods in a single repository including PLR \citep{jiang2021prioritized}, PAIRED \citep{dennis2020emergent, mediratta2023stabilizing}, Robust PLR, REPAIRED \citep{jiang2021replay}, and ACCEL \citep{parker2022evolving}. Watts is another curriculum learning library focused on open-endedness with implementations of POET and PAIRED \citep{dharna2022watts}. It atomizes components of the open-ended framework into modules to create novel methods by combining these components in new ways. Their work also compares these algorithms on a suite of common benchmarks. \citet{yuan_roger2024rlexplore} introduce a library of instrinsic motivation algorithms called RLeXplore, and use their library to identify algorithmic components of PPO and SAC \citep{haarnoja2018soft} that impact the performance of popular exploration bonuses. TeachMyAgent \citep{romac2021teachmyagent} is a complete benchmarking library for CL, which includes two procedural Box2D \citep{brockman2016openai} environments, a collection of ACL methods, and several student algorithm implementations. Their work categorizes several challenges for ACL methods including task space feasibility, robustness to novel RL algorithms, and required expert knowledge. TeachMyAgent implements several older ACL methods which \citep{jiang2021replay} showed underperform relative to UED baselines.

Recently, JAX has become a popular choice for writing fast deep learning libraries that can be run end-to-end on hardware accelerators \citep{jax2018github}. Minimax \citep{jiang2023minimax} provides JAX-based implementation of the UED algorithms in the DCD library, leading to significantly faster training. JaxUED \citep{coward2024jaxued} refactors the UED algorithms in Minimax into single-file implementations for faster prototyping, inspired by CleanRL \citep{huang2022cleanrl}. Both of these libraries are intended to be used with JAX-based environments, which allow you to run the entire training loop on hardware accelerators. However, they provide only moderate performance benefits to the complex CPU-based games that motivate open-endedness research, at the cost of inconvenient code constraints. These libraries also suffer from the same lack of portability as previous curriculum learning systems -- their algorithms can not be easily applied to RL code written in any other library.

Syllabus distinguishes itself from previous works by making minimal assumptions on the training code, rather than providing its own training system that intermingles RL and CL. By defining a simple, uniform API for interfacing with a curriculum, Syllabus makes it possible to add these algorithms to nearly any RL system with minimal code changes. An additional benefit of this approach is that all of the algorithmic details are contained in a single place, rather than distributed across the codebase. Syllabus also provides the only general-purpose infrastructure for synchronizing curricula across CPU-based environments in multiple processes. Curriculum learning algorithms almost exclusively interact with environments rather than policy optimization, and Syllabus takes advantage of this through a unique paradigm of portable infrastructure.

\section{Design Philosophy}
\label{sec:design}

Syllabus aims to simplify the process of developing curriculum learning methods, combining them with RL algorithms, and applying them to challenging domains. As such, it is built to be compatible with as many different RL libraries and multiprocessing methods as possible. These goals motivate the following key points of our design philosophy:

\begin{enumerate}
    \item Syllabus should be agnostic to the choice of reinforcement learning framework.
    \item Syllabus should have general APIs that support any form of curriculum learning.
    \item Integrating Syllabus into training infrastructure should require minimal code changes.
    \item Code complexity should scale with the complexity of the curriculum learning algorithm.
    \item Algorithm implementations should be contained in a single file.
\end{enumerate}

The first point motivates many of the implementation choices in Syllabus that may seem odd in isolation. To maintain compatibility with several libraries we must honor the Gymnasium and PettingZoo environment APIs \citep{brockman2016openai, towers2024gymnasium, terry2021pettingzoo} and write systems that work with many different multiprocessing solutions. For instance, Gymnasium, Stable Baselines 3 \citep{stable-baselines3}, and RLLib \citep{liang2018rllib} all provide their own vector environment implementations.

Syllabus's utility is tied to its ability to integrate new forms of curriculum learning. The diversity of existing CL algorithms is highlighted in \autoref{sec:introduction} and \autoref{sec:curricula}. Supporting all of these methods in a single API necessitates some complexity. When one method requires a new interface, we prefer to have a heterogeneous, modular API rather than complicate the interface for all methods. For example, we define a separate Curriculum and Agent API for task-based and opponent-based curricula respectively. These components are described in \autoref{sec:curricula} and can be used separately or combined to form joint curricula over tasks and opponents. In each case, we create the smallest possible API to minimize complexity for users.

Our focus on single-file implementations is inspired by the success of CleanRL \citep{huang2022cleanrl}. CleanRL provides single-file implementations of popular RL algorithms to support fast iteration and transparency, making it easy for researchers to reliably report algorithmic details. Though end-to-end single-file training scripts are inherently non-portable, Syllabus encapsulates all CL logic for each algorithm in a single class to capture much of the same transparency and simplicity.

\section{Syllabus APIs}
\label{sec:API}

\begin{figure}[t!]
\centering
  \includegraphics[width=0.9\textwidth]{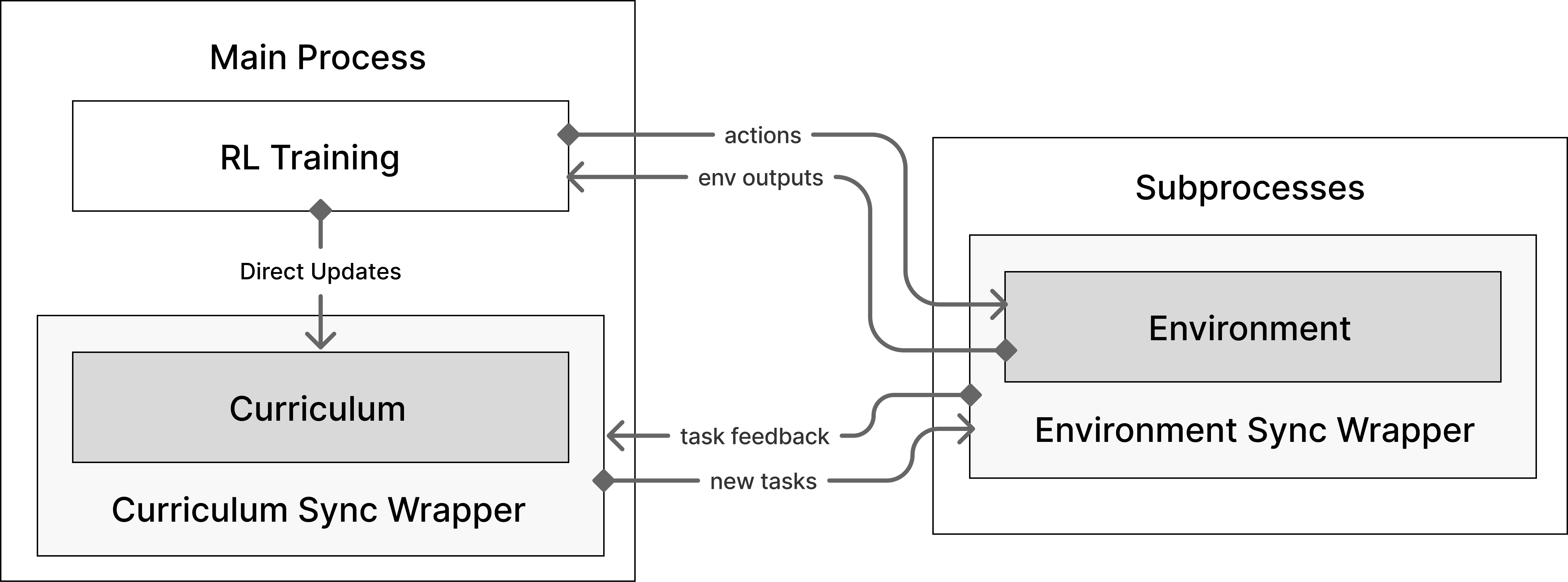}
  \caption{Syllabus with a standard asynchronous RL training setup.}
  \label{fig:multiprocessing}
\end{figure}

Syllabus designates responsibility for maintaining sampling distributions over the task space to a \texttt{Curriculum} class, and implements task swapping through a wrapper over the environment. This provides a uniform interface for setting the current task of an environment, which we explain fully in \autoref{app:task_api}. Each environment also defines a \texttt{TaskSpace} that represents the full range of tasks that can be used to train agents. Each API is designed to be simple to use and to support future use cases.

\subsection{Curriculum API}
\label{sec:curricula}
In Syllabus, a \texttt{Curriculum} is responsible for maintaining a distribution over the task space and implementing a sampling function for selecting tasks. Automatic curriculum learning methods use feedback from the RL training process to update their sampling distribution. The Curriculum API provides multiple options for incorporating information from the environments and policy, either manually or automatically through Syllabus's synchronization infrastructure. The multiprocessing approach is explained more thoroughly in \autoref{sec:multi}. The curriculum can be updated after each step, episode, or completed task with the \texttt{update\_on\_step}, \texttt{update\_on\_episode}, or \texttt{update\_task\_progress} methods respectively. These update options allow us to implement a diverse range of curriculum learning algorithms under the same API, as we show in \autoref{sec:implementations}.

\subsection{Agent API}
 The Agent API is a subset of the Curriculum API that defines curricula over co-players in multi-agent games. These algorithms store co-players and load them at a later time based on some sampling criteria. The currently implemented algorithms are forms of self-play (SP) \citep{5392560, 10.1145/203330.203343} where the opponent is a previous copy of the online policy. These are intended for two-player competitive games, but the API supports many-agent general-sum games.

\begin{enumerate}
    \item \textbf{Fictitious Self Play (FSP)} - maintains a history of past copies of the agent as opponents \citep{10.5555/3045118.3045205}. FSP uniformly samples an opponent from history to prevent strategic cycles.
    \item \textbf{Prioritized Fictitious Self Play (PFSP)} - like FSP, PFSP maintains a history of past opponents to sample during training. PFSP selects the opponent with the highest win-rate against the current agent. This prevents the curriculum from spending a disproportionate amount of time training against opponents that the agent already performs well against.

\end{enumerate}

Syllabus supports the simultaneous use of opponent-based (e.g. FSP, PFSP) and task-based (e.g. DR, PLR) curricula through the \texttt{DualCurriculumWrapper}. This wrapper extends the Curriculum API, allowing the user to sample from a joint task space and update both curricula at once. This API allows users to experiment with different joint curricula without changing their training code.

\subsection{Task Space API}
Defining a task space for training agents is a fundamental challenge in curriculum learning, similar to reward shaping for policy optimization. Task spaces are domain-specific and must be carefully designed for each environment to create an effective curricula. In most benchmark environments, tasks exist in a low-dimensional discrete or continuous space, while more complex environments may use a combination of discrete and continuous variables or intricate predicate systems, as seen in XLand \citep{team2023human, nikulin2023xlandminigrid} and Neural MMO \citep{suarez2019neural}. Curriculum learning implementations are often restricted to the task space representations from the environments that they were originally designed for, but Syllabus's Task API removes these limitations, making it possible to explore new task spaces without modifying algorithm code.

The Task Space API simplifies task space definition and curriculum compatibility by mapping tasks into a Gym Space \citep{brockman2016openai}. This allows users to define tasks in a format suited to their environment while maintaining a simple representation within the curriculum code. For example, while Prioritized Level Replay originally used level seeds as tasks, our implementation supports any discrete task list. The PLR code handles tasks as integers for streamlined algorithm logic, while the environment interprets them as seeds, map encodings, reward functions, etc. This approach also reduces bandwidth for task transfer between processes and enables task indexing for defining separate training and validation sets.

\begin{figure}[!t]
\begin{minted}[frame=single, linenos]{python}
    curriculum = DomainRandomization(task_space)
    curriculum = make_multiprocessing_curriculum(curriculum)

    env = NetHackScore()
    env = NetHackTaskWrapper(env)
    env = GymnasiumSyncWrapper(env, curriculum.components)
\end{minted}

\caption{Using Syllabus for curriculum learning with just a few lines of code.}
\label{fig:code-sample}
\end{figure}

\subsection{Multiprocessing Infrastructure}
\label{sec:multi}
The real practical challenges in curriculum learning come from synchronizing curricula using feedback from environments running in multiple processes. Syllabus's infrastructure is designed to separate curriculum and multiprocessing logic to provide interoperability with many different forms of asynchronous RL infrastructure. It uses a bidirectional sender-receiver model in which the curriculum sends tasks and the environment sends feedback from playing the provided task. The curriculum synchronization wrapper adds multiprocessing functionality to a \texttt{Curriculum} and an environment synchronization wrapper adds the same functionality to the environment without changing the Gym interface. The environment synchronization wrapper automatically sends feedback to the curriculum after each step, episode, or completed task depending on the curriculum method. You can also update the curriculum with training metrics directly from the main learner process. \autoref{fig:multiprocessing} shows a diagram of how these components interconnect. The curriculum does not need immediate feedback or block training, so all updates are batched to reduce multiprocessing overhead, and task sampling is can be buffered to prevent delays at the start of each episode. Crucially, adding Syllabus's synchronization to existing RL training code requires only a few lines of code, shown in \autoref{fig:code-sample}. 

The user-facing curriculum and environment code follows our design goals stated in \autoref{sec:design}, while the multiprocessing infrastructure is engineered to ensure stability and reduce the risk of bugs. To guarantee that researchers will not need to spend time reading or debugging this code, Syllabus includes thorough integration tests, smoke tests, regression tests, and optimization benchmarks for all multiprocessing code, tested with all of the implemented curriculum learning methods. More details and performance numbers can be found in \autoref{app:testing} and \autoref{app:optimization}.

\subsection{Automatic Curriculum Learning Implementations}
\label{sec:implementations}
Syllabus includes portable implementations of several popular curriculum learning baselines. We prioritize modern methods which have been successfully applied to complex environments, and plan to add many more algorithms in future versions of Syllabus. We also provide utilities for manual curriculum learning or transfer learning. For example, we implement Simulated Annealing \citep{doi:10.1126/science.220.4598.671}, an expanding sampling range curriculum, and a sequential curriculum. The sequentual curricula trains agents on a list of individual tasks or entire curricula in stages that run until a predetermined number of steps or user-defined stopping conditions are met. Supporting all of these diverse methods helps to demonstrate the generality of Syllabus's interfaces.

\textbf{Prioritized Level Replay (PLR)} - a popular UED method which maintains a buffer of levels with high learning potential \citep{jiang2021prioritized}. PLR typically prioritizes tasks with a high value loss, though recent work explores other options \citep{jiang2021replay, jackson2024discovering}. It also tracks the staleness of each task's value loss score, and assigns some probability to very stale tasks.

We generalize this implementation to support arbitrary task spaces instead of only environment seeds, and implement an asynchronous version that can be updated with arbitrary batch sizes. We also implement Robust PLR \citep{jiang2021replay} as an initialization option for PLR. PLR is a core component of several more recent UED methods \citep{parker2022evolving, samvelyan2022maestro} and has been successfully applied to complex domains \citep{team2023human}. We sample tasks based on the L1 value loss as defined below, where $\lambda$ and $\gamma$ are the GAE and MDP discount factors respectively, and $\delta t$ is the TD-error at timestep $t$:

\begin{equation}
\centering
\label{eq:pvl}
\frac{1}{T}\sum_{t=0}^{T} \left|\sum_{k=t}^T(\gamma\lambda)^{k-t}\delta_k\right| 
\end{equation}

\textbf{Learning Progress (LP)} - this method was developed by \citet{kanitscheider2021multi} to train PPO agents in Minecraft. Using this algorithm and an exploration bonus, their agents were able to consistently acquire diamonds, drastically surpassing PPO alone. They define a set of pass/fail tasks for the agent to achieve, and prioritize tasks with a recent change in success rate, which they call learning progress. This method tracks a fast and slow exponential moving average (EMA) of the agent's success rate for each task, which is evaluated periodically throughout training. Learning progress is calculated by taking the absolute difference between these EMA values and performing several normalization steps. High learning progress indicates a task that the agent is beginning to learn or forget, so those tasks are prioritized during sampling.

\textbf{Open-endedness via Models of human Notions of Interestingness (OMNI)} - this is an extension to the Learning Progress curriculum introduced by \citep{zhang2023omni} which asks an LLM to identify "interesting" tasks given the agent's current success rates on each task. The LLM filters out tasks that are uninteresting given high proficiency at another task. Starting with the highest success rate, the LLM partitions tasks into interesting and uninteresting sets, and masks uninteresting tasks out of the sampling distribution generated by the LP curriculum. This approach has recently been extended to allow the LLM to generate new interesting tasks defined in code \citep{faldor2024omni}.

\textbf{Sampling for Learnability (SFL)} - this method is a hybrid of the approaches used by PLR and LP. It uses a simple metric for prioritizing tasks by sampling according to $p(1-p)$ where $p$ is the task success rate, a metric called learnability that was originally described in \citep{tzannetos2023proximal}. The method as introduced by \citet{rutherfordno} samples from a mixture of a uniform distribution over the top $k$ most learnable tasks and a uniform distribution over all tasks. The mixing ratio of these distributions is controlled by a hyperparameter $p$. We additionally implement a version of this method that samples directly from the full distribution generated by $p(1-p)$. We find that it sometimes outperforms SFL while removing two environment-dependent hyperparameters.


\begin{figure}[t!]
  \begin{subfigure}[t]{0.48\textwidth}
      \centering
      \includegraphics[width=1.0\textwidth]{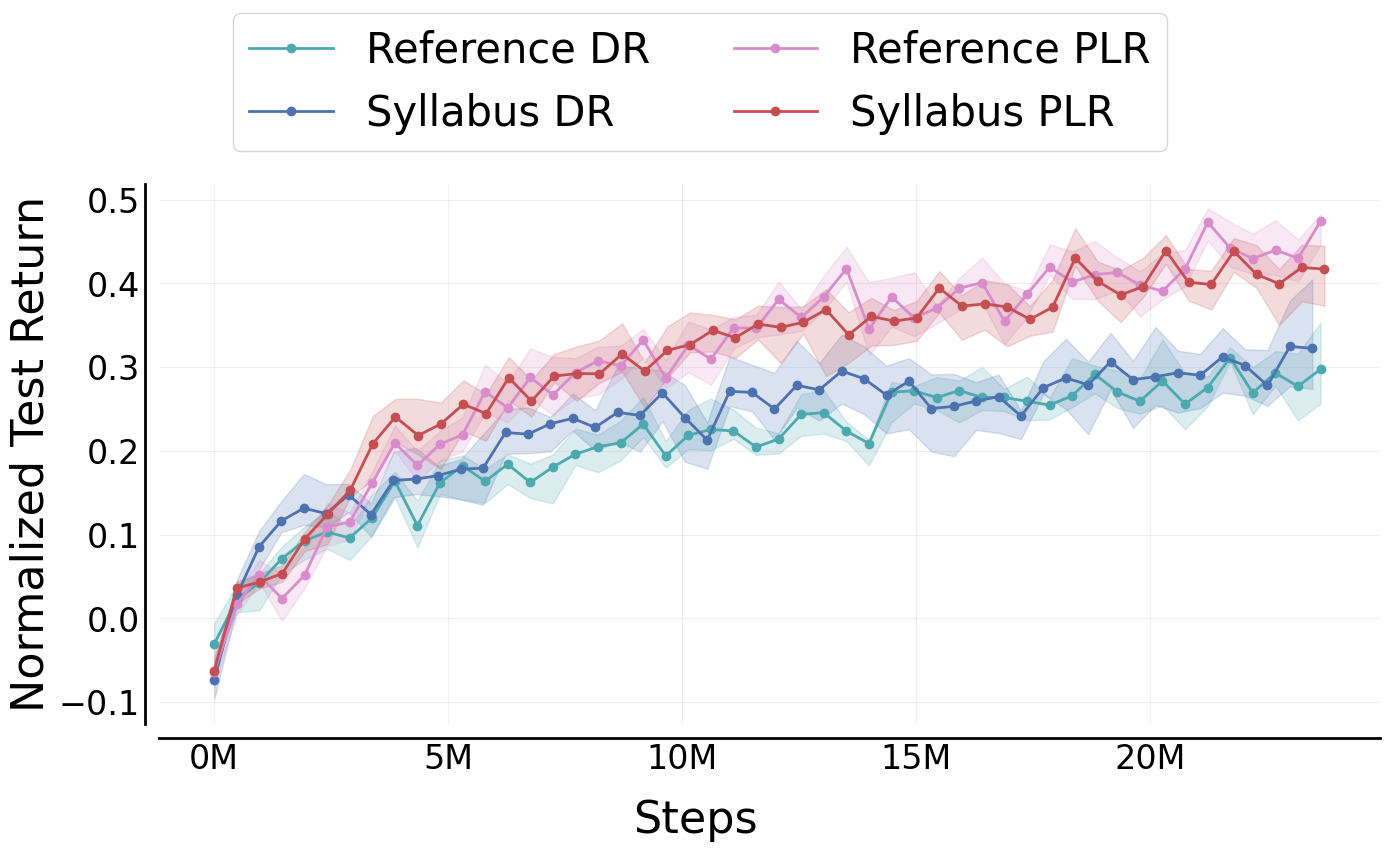}
      \caption{}
      \label{fig:procgen_reference}
  \end{subfigure}
  \begin{subfigure}[t]{0.48\textwidth}
      \centering
      \includegraphics[width=1.0\textwidth]{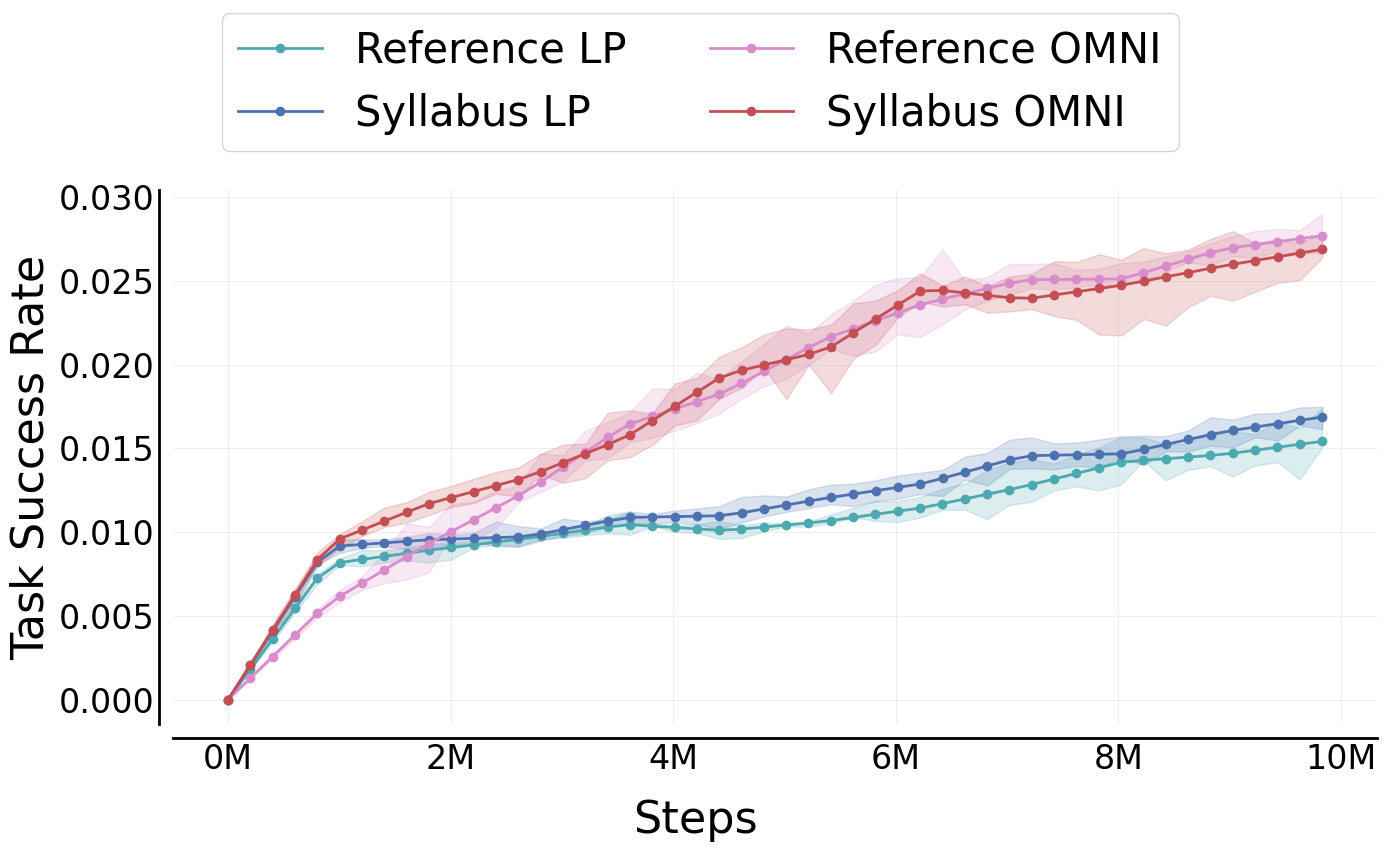}
      \caption{}
      \label{fig:crafter_reference}
  \end{subfigure}
  \begin{subfigure}[t]{1.0\textwidth}
    \includegraphics[width=1.0\textwidth]{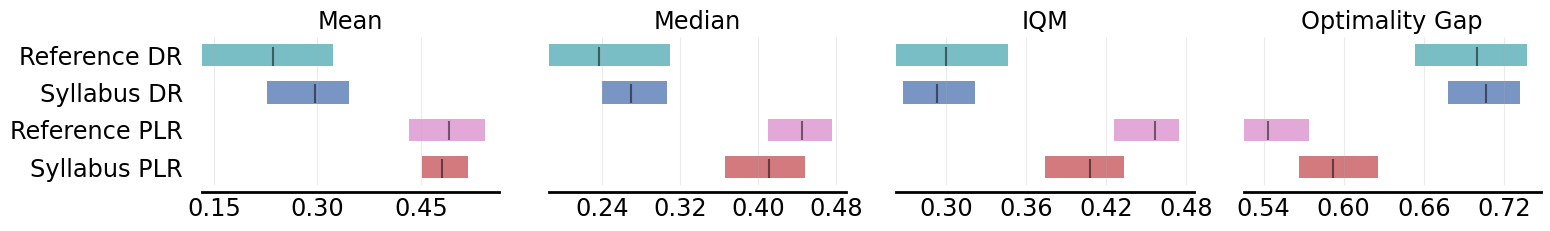}
    \caption{}
  \end{subfigure}
  \begin{subfigure}[t]{1.0\textwidth}
    \includegraphics[width=1.0\textwidth]{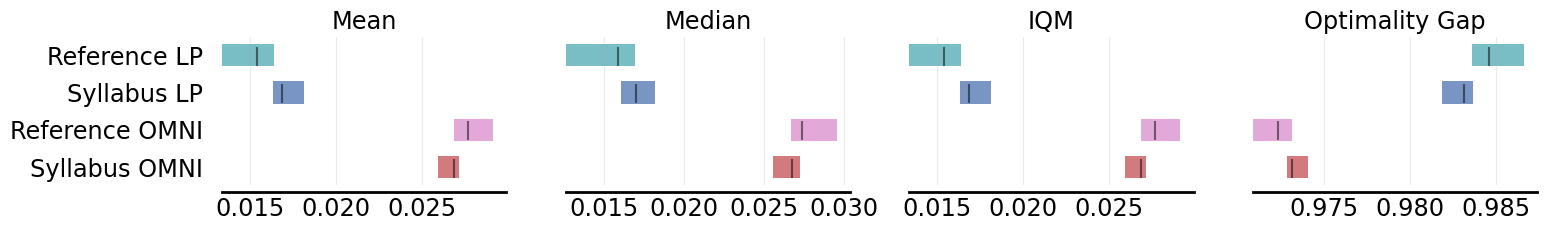}
    \caption{}
  \end{subfigure}

  \caption{\textbf{(a)} Mean normalized test returns for Syllabus's implementation vs. the original implementation of Prioritized Level Replay from \citep{jiang2021prioritized} on 10 Procgen environments. Domain Randomization is also included for reference. \textbf{(b)} Mean task success rate for Syllabus's Learning Progress and OMNI implementations vs. the reference implementations from \citep{zhang2023omni} on Crafter. \textbf{(c)} and \textbf{(d)} 95\% Stratified bootstrap confidence intervals for Procgen and Crafter.}
  \label{fig:reference}
\end{figure}

\section{Reproduction Experiments}
In order to demonstrate the correctness of Syllabus's infrastructure and algorithms, we reproduce the experiments for each automatic curriculum learning method from the paper in which they were introduced. We test PLR and DR on Procgen as in \citep{jiang2021prioritized}, and we test LP and OMNI on Crafter \citep{hafner2021benchmarking} as in \citep{zhang2023omni}. Syllabus makes it easy to test our implementations; we simply take any open-source reference codebase, disable the curriculum, and replace it with a few lines of Syllabus code. This lets us quickly evaluate new ACL implementations without recreating experiments in a new RL library, thereby minimizing methodological errors.

Procgen is a collection of 16 procedurally generated games designed to test sample efficiency and generalization \citep{cobbe2020leveraging}. Procgen levels are generated from a seed, so we use a task space of 200 training seeds for all curricula. Our experiments focus on a subset of 10 Procgen environments, but otherwise use the same methodology as \citet{jiang2021prioritized} and \citet{cobbe2020leveraging}. Like these works, we normalize test returns by dividing returns by the empirical return range in each Procgen environment. Our full methodology is explained more thoroughly in \autoref{app:procgen}. Crafter is a procedurally generated, grid-world environment modeled after Minecraft \citep{guss2019minerl, hafner2021benchmarking}. Agents collect resources, craft tools, and fight monsters to survive. The environment assigns tasks to the agent such as "collect 5 coal" or "make 1 iron pickaxe", which the agent receives 1 reward for completing. The curricula in \citet{zhang2023omni} are therefore curricula over reward functions, based on the agent's current competence on each task.

We show in \autoref{fig:reference} that the normalized test returns for PLR and DR on Procgen over the course of training almost precisely match the reference implementations from \citet{jiang2021prioritized}. Similarly in Crafter, we see that our implementations of LP and OMNI achieve the exact same task success rates throughout training as the reference implementations. As in \citep{zhang2023omni} the task space includes 15 main tasks such as "collect 1 coal", 90 repeat tasks such as "collect 8 coal", and 1024 impossible tasks that always have 0 success rate. We evaluate the success rates for each of the 105 possible tasks for 4 episodes each over the course of training. These experiments provide strong evidence that our multiprocessing infrastructure and algorithm implementations are correct. The full details of our methodology for these experiments can be found in \autoref{app:procgen} and \autoref{app:crafter}. We also use the Open RL Benchmark tools \citep{huang2024open} to plot stratified bootstrap confidence intervals for the mean, median, and interquartile mean of each method as recommended by \citet{agarwal2021deep}.

\autoref{fig:reference} shows that our implementations of PLR, LP, and OMNI match the performance of the reference implementations. Syllabus's portable design means that we can apply these algorithms to new domains with the confidence that the curriculum learning portion of the code is completely correct. In \autoref{sec:new-baselines} we demonstrate this by applying ACL to 2 new domains which have not previously been studied with curriculum learning.

\section{New Baselines}
\label{sec:new-baselines}

We demonstrate Syllabus's versatility by applying ACL to two new complex domains, Neural MMO and NetHack, with baselines implemented in specialized RL libraries. We also present new baselines for each ACL method on Procgen and Crafter, trained using CleanRL \citep{huang2022cleanrl} and TorchAC respectively. This is the first direct comparison using a shared benchmark of the learning progress methods LP and OMNI against UED methods like PLR and SFL.

\subsection{Neural MMO 2.0 in PufferLib}

Neural MMO 2.0 is a complex multi-agent simulation inspired by massively multiplayer online games \citep{suarez2019neural, suarez2024neural}. Agents can collect resources, learn skills, trade goods, and fight non-player characters or other agents. It has a predicate task space which allows users to define objectives in Python code. The baseline for the 2023 Neural MMO competition was written in PufferLib because it supports complex action spaces and multi-agent environments where agents can die mid-episode, which complicates learning code \citep{terry2021pettingzoo, suarez2024neural}. We show that Syllabus can be used in this environment with 128 agents and a massive task space. \\

All of the automatic curriculum learning methods in Syllabus (excluding FSP and PFSP) were designed for single-agent environments. To apply single-agent curricula to a multi-agent environment, we update our curricula with feedback from all 128 agents controlled by our self-play policy. Each agent's experience is treated as trajectories from separate single-agent environments. We do not test FSP or PFSP because Neural MMO is a mixed competitive-cooperative game, not purely zero-sum.

\subsection{NetHack in Moolib}

NetHack is a popular text-based dungeon-crawler released in 1987, and adapted into an RL environment by \citet{kuttler2020nethack}. It's a complex, procedurally generated game in which winning or "ascending" can take more than 50,000 steps for human players. Ascending requires players to solve puzzles using common sense, knowledge of mythology, and game-specific tricks, all while collecting equipment, fighting monsters, and scavenging for food. NetHack remains one of the hardest benchmarks for online RL methods, which lag behind hand-crafted symbolic agents and behavior cloning baselines \citep{kuttler2020nethack, tuyls2023scaling, piterbarg2024nethack}.

Importantly, NetHack can be simulated extremely quickly, shifting the training bottleneck from data collection to policy optimization and inference. To address this, NetHack baselines use specialized training libraries like TorchBeast \citep{kuttler2019torchbeast}, Sample Factory \citep{petrenko2020sample}, and moolib \citep{moolib2022}. Sample Factory and moolib use asynchronous PPO (APPO), which creates separate copies of the policy for action inference and optimization. Moolib spawns servers for policies and environments that communicate via remote procedure calls, allowing it to scale to many machines. We demonstrate that Syllabus is easy to use with moolib in a single GPU setting. \\

\subsection{Methodology}
\label{sec:methodology}

\begin{figure}[t!]
  \begin{minipage}[t]{0.48\textwidth}
      \centering
      \includegraphics[width=1.0\textwidth]{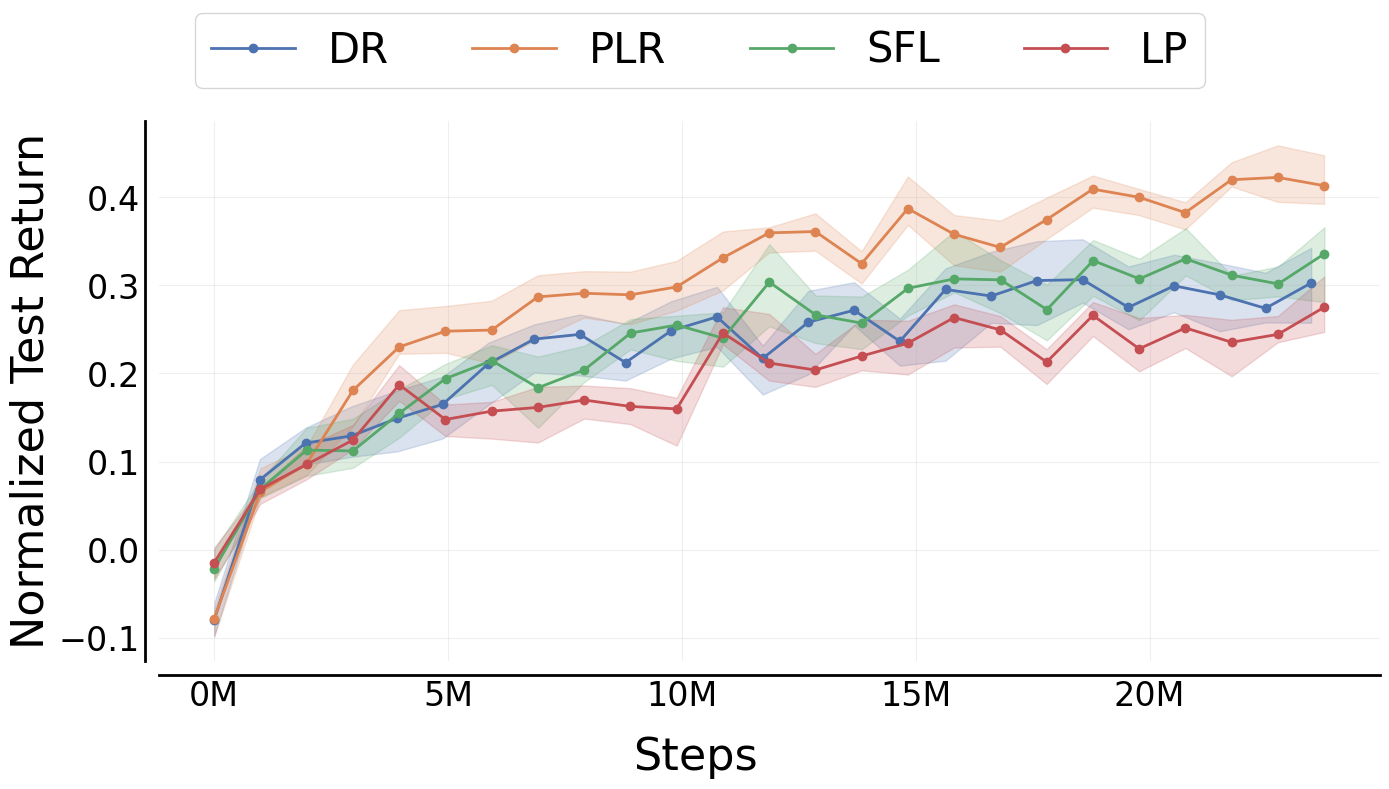}
      \subcaption{Procgen}
      \label{fig:procgen}
  \end{minipage}
  \begin{minipage}[t]{0.48\textwidth}
      \centering
      \includegraphics[width=1.0\textwidth]{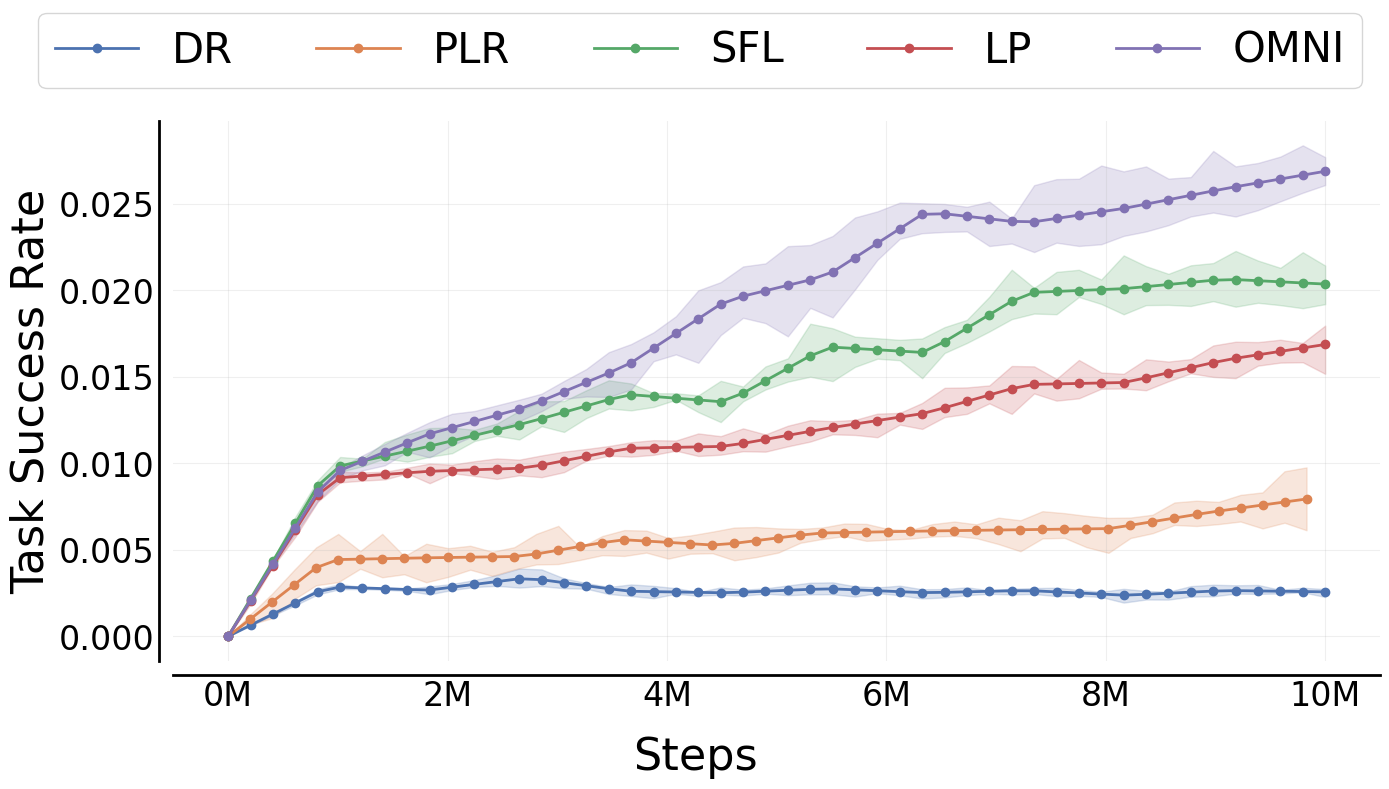}
      \subcaption{Crafter}
      \label{fig:crafter}
  \end{minipage}
  \begin{minipage}[t]{0.48\textwidth}
      \centering
      \includegraphics[width=1.0\textwidth]{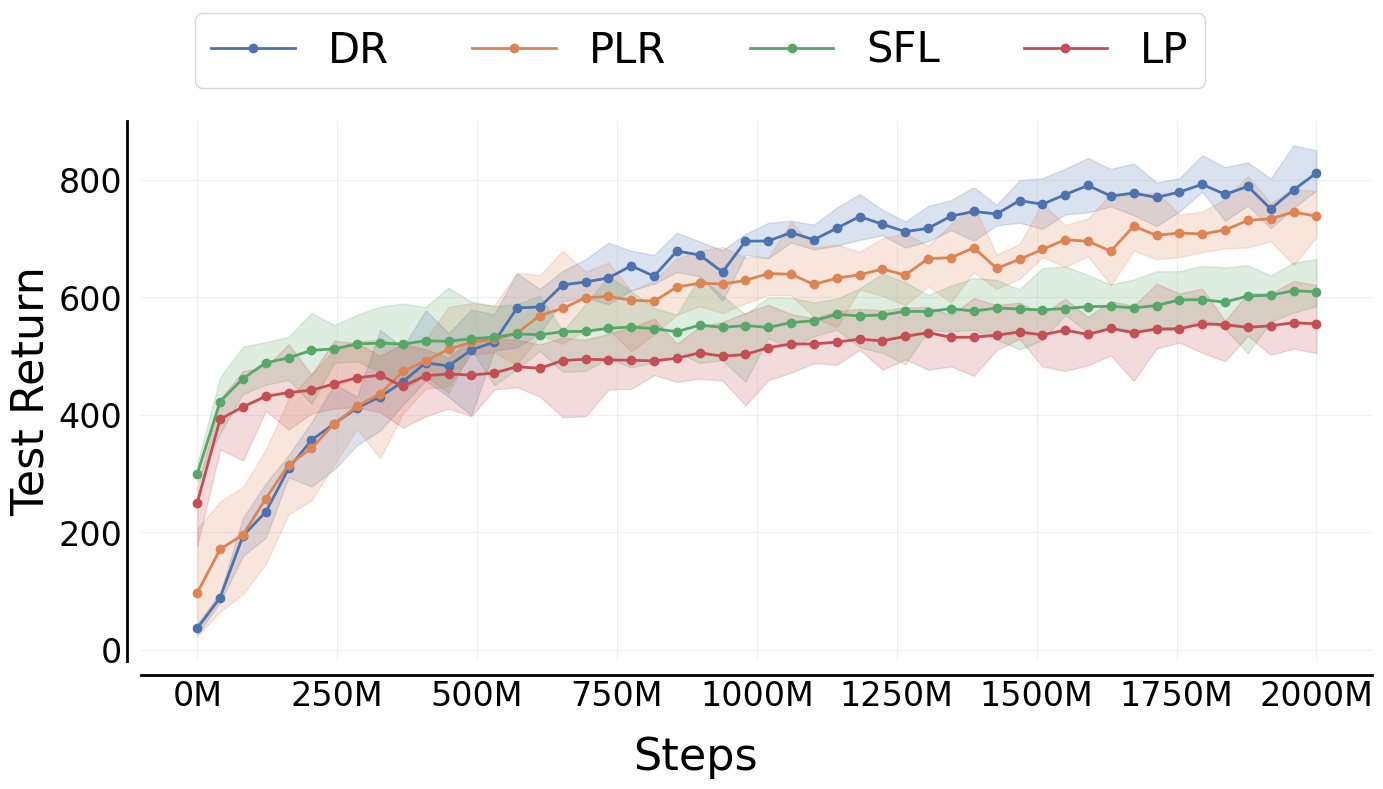}
      \subcaption{NetHack}
      \label{fig:nethack}
  \end{minipage}
  \hspace{0.04\textwidth}
  \begin{minipage}[t]{0.48\textwidth}
      \centering
      \includegraphics[width=1.0\textwidth]{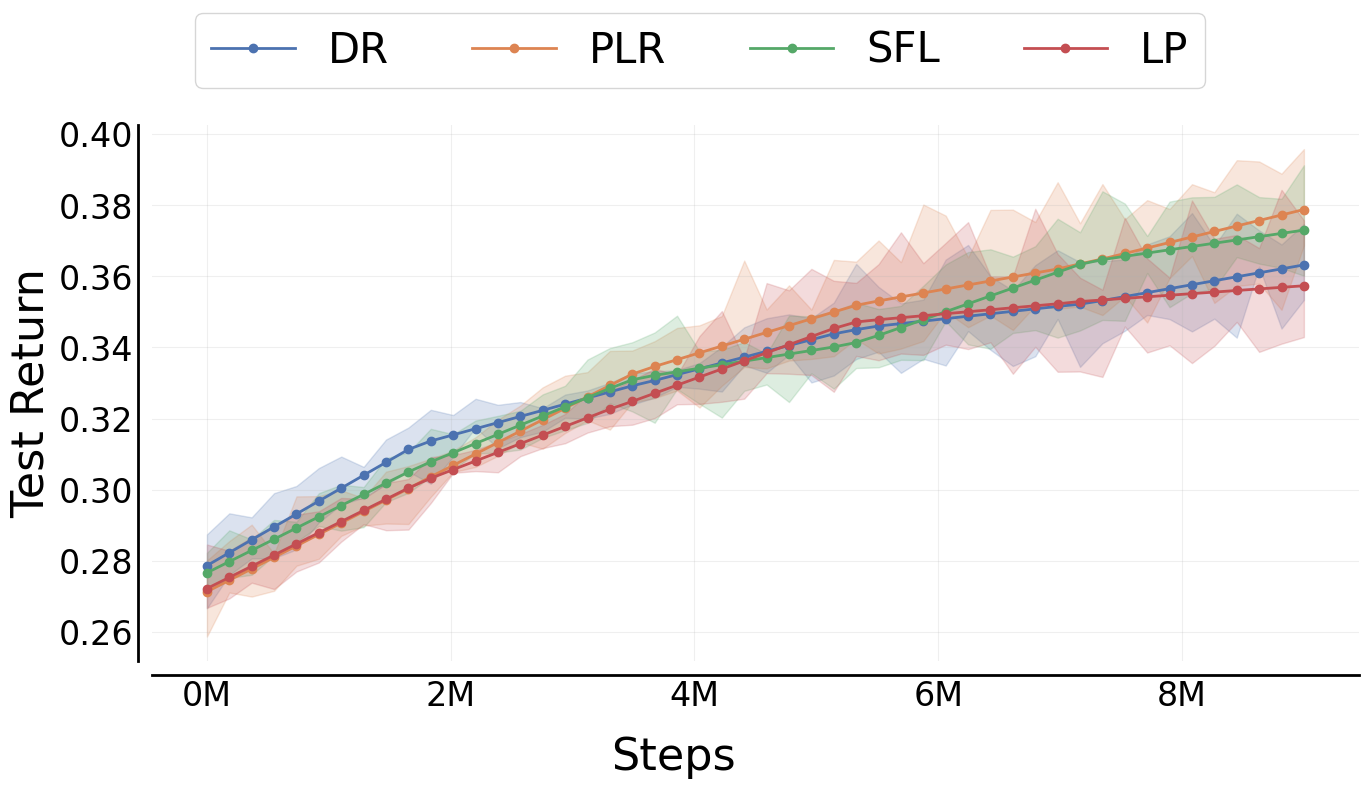}
      \subcaption{Neural MMO}
      \label{fig:nmmo}
  \end{minipage}
  \caption{Automatic curriculum learning results on \textbf{(a)} Procgen, \textbf{(b)} Crafter, \textbf{(c)} NetHack, and \textbf{(d)} Neural MMO. 95\% Stratified bootstrap confidence intervals can be found in \autoref{app:experiments}}
  \label{fig:results}
\end{figure}

LP and SFL rely on a task success metric to evaluate seeds, which is not a standard feature of RL environments. For Crafter, these are provided via binary pass/fail achievements. For all other environments, we define the success metric as the scaled, clipped, episodic return achieved by the agent. The exact scaling was manually selected based on the environment's typical reward range and is described in \autoref{app:experiments}. OMNI is only evaluated on Crafter because it is not possible to identify "interesting" level seeds or maps by number alone, and there is no obvious way to describe arbitrary environment instantiations to an LLM. We use a grid search to tune every curriculum's hyperparameters in each environment, which we explain more thoroughly in \autoref{app:hyperparameters}.

Each of the libraries used in this paper have different design philosophies, software architectures, and multiprocessing implementations. Despite this, Syllabus can easily work with all of them with a few lines of library-agnostic code. We have an additional example in \autoref{app:cartpole-rllib} of training CartPole agents in RLLib \citep{liang2018rllib} to demonstrate how Syllabus can be used with Ray multiprocessing. We also experiment with adding PLR to Phasic Policy Gradients \citep{cobbe2021phasic}, an extension to PPO, in \autoref{app:ppg} to show that Syllabus is not limited to augmenting PPO. More details for all of our experiments can be found in \autoref{app:experiments}, and the code and data are publicly available on GitHub and Weights \& Biases respectively.

\section{Results}
\label{sec:results}

We find that only PLR outperforms DR when applied to Procgen's level seeds. In this setting, LP underperforms and SFL, matching the simpler and cheaper DR. Neither of these newly evaluated methods perform well on Procgen despite hyperparameter tuning. We plot individual training curves for each Procgen environment in \autoref{fig:procgen_all} and find that there are no environments where LP or SFL outperform PLR, but there are some environments where they outperform DR. Also of note, DR is the only method which achieves nonzero return on Dodgeball. On our Crafter environment where 90\% of tasks are impossible, the DR agent fails to learn a reasonable policy, barely outperforming and random policy. PLR is able to identify meaningful tasks and allows the agent to slowly improve its competence. \citet{zhang2023omni} demonstrated the effectiveness of LP and OMNI on Crafter, but we find that SFL performs significantly better than LP. As a result, it seems natural to apply OMNI's LLM-based interestingness filtering on top of SFL instead of LP, but we found that this does not work well in practice \autoref{app:omni-learnability}. These results may suggest that task success rates are a more effective prioritization metric than value predictions when available. In their absence, value predictions might be a better approximation of competence than our return-based success metric defined in \autoref{sec:methodology}. We leave a thorough investigation of this inconsistency as future work. 

Crafter stands out as an effective testbed for CL research. Unlike other environments, it is easy to see which methods outperform others in \autoref{fig:crafter}. Crafter and Minecraft are designed such that proficiency on one task immediately makes the next task learnable, because the agent is guaranteed to have the knowledge and tools to acquire the next item. We suspect that this reward-based, linear task space highlights the strengths of CL algorithms. However, that nice structure is not easy to identify in most problems, so it is not a replacement for evaluating on complex environments.

In NetHack, we see that PLR performs similarly to DR. This is unsurprising because a single seed of NetHack can diverge significantly in just a few steps, so value predictions have extremely high variance and level seeds have little control over the environment instance's difficulty. However, we also see that LP and SFL accelerate learning at the start of training, but lead to lower asymptotic performance. As we saw in Crafter, success rates may be a more effective way of identifying learnable seeds, but prioritizing these seeds is not a good long-term strategy for boosting performance in NetHack. No curriculum provides any asymptotic benefits over DR in NetHack. 

We also see in \autoref{fig:nmmo} that none of our automatic curriculum learning methods for selecting maps in Neural MMO perform well compared to DR. In complex, many-agent environments the map may have minimal influence over the difficulty of the environment, suggesting that curricula over different dimensions could be more effective. Additional experiments using a manually designed curriculum of progressively harder reward functions can be found in \autoref{app:experiments}, which we find outperforms the default survival reward on several metrics.

Overall, we find that most curricula perform poorly outside of the environment in which they were originally tested. Agents in both Procgen and Crafter have been shown to benefit from CL using PLR and LP respectively \citep{jiang2021prioritized, zhang2023omni}, yet neither of these methods perform well when applied to the opposite environment. SFL is one of the few successful evaluations scoring higher than LP on Crafter and at least matching DR on Procgen, despite being developed on simpler JAX-based environments. None of the methods provide any noticeable improvement over the DR baseline in Neural MMO or NetHack. These results suggest that curricula over initial environment conditions are not effective in complex, long-horizon games, and that we may need to develop more sophisticated methods for these environments.

\section{Discussion and Future Work}
\label{sec:discussion}
Multi-domain research, focusing on challenging, long-horizon environments, is crucial to developing generalizable and robust RL methods. \citet{schaul2011measuring}, \citet{bellemare2013arcade}, and \citet{Castro_ALE} have argued that video game environments are an ideal benchmark for AI agents because they are designed for human players. This makes them inherently difficult yet learnable while being compelling and interpretable. In practice, solving these complex problems with reinforcement learning often requires custom infrastructure. NetHack agents are trained in libraries that can take advantage of its fast simulation speed. Neural MMO agents are built with PufferLib because it natively supports the complexities of multi-agent interactions. Syllabus is designed for this reality of training agents in complex environments. Separating curriculum learning from RL code allows us to apply the same methods to new environments without re-implementation. We hope this will improve the reproducibility of curriculum learning research and help to push the field away from toy environments and toward domains that challenge modern RL methods.

Much of the recent UED infrastructure has been written in JAX. This allows environments and training code to be parallelized on hardware accelerators, producing experimental results hundreds of times faster than equivalent CPU-based environments. JAX is a powerful tool for conducting fast research, but it also enforces strict requirements on how code is written, slowing development and incentivizing simplistic environments. It will be challenging for JAX-based simulations to reach the complexity of CPU-based research environments, much less the professionally developed video games that have historically been benchmarks for reinforcement learning. Syllabus is designed to extend the benefits of curriculum learning to more complex domains.

Our results suggest that although curriculum learning is effective in moderately challenging single-agent environments like Procgen and Crafter, standard approaches may not be sufficient for complex, long-horizon, and multi-agent environments like NetHack and Neural MMO. These results could likely be improved by using curricula over more meaningful axes of the task space, such as reward functions in NetHack or opponent strategies in Neural MMO. However, more complex task spaces also introduce new challenges. For instance, training on multiple reward functions will require careful per-task reward normalization. One practical finding from our experiments is that SFL performs at least as well as LP in each environment, and only scores lower than DR in NetHack. In addition, our full distribution variant of SFL performs comparably to the original top K implementation and requires no hyperparameter tuning. Therefore, our full distribution implementation of SFL is a strong initial choice for exploring curriculum learning on new domains.

Syllabus is actively evolving as we add features, benchmark new methods, and expand support for general curriculum learning approaches. We hope it enables future research and encourages evaluation on more challenging domains. We hope it serves as a foundation for future curriculum learning research, particularly in challenging environments.

\section{Limitations}
\label{app:limitations}

Syllabus defines a separate multiprocessing pathway to send data to the curriculum. When curricula require standard environment outputs, this will send the same information as the RL training pathway, potentially leading to bandwidth or processing bottlenecks. It is possible to avoid this duplication by writing additional code to forward data from the training process to the curriculum. Syllabus does not support distributed environment servers, though it is possible via RPC calls with minimal modification. Syllabus's multiprocessing infrastructure cannot automatically communicate with environments vectorized in JAX or C without additional code, though the curricula can still be used directly. As a result, it can not be used in fully parallel training systems. Syllabus does not currently include methods that train additional networks or create new reward components. However, using those methods alongside Syllabus will be no harder than using them without Syllabus. 

\section{Acknowledgments}

We would like to thank Minqi Jiang for his insightful design feedback, for his open-source UED implementations, and for helping to debug the PLR baselines. We also thank Jenny Zhang for assisting our implementation of LP and OMNI based on her open-source code base. We would also like to thank Joseph Suárez for his valuable software architecture suggestions. Finally, we thank the reviewers for their thoughtful consideration of our work and helpful suggestions.

\bibliography{bibliography}

\begin{thebibliography}{106}
\providecommand{\natexlab}[1]{#1}
\providecommand{\url}[1]{\texttt{#1}}
\expandafter\ifx\csname urlstyle\endcsname\relax
  \providecommand{\doi}[1]{DOI: #1}\else
  \providecommand{\doi}{DOI: \begingroup \urlstyle{rm}\Url}\fi

\bibitem[Agarwal et~al.(2021)Agarwal, Schwarzer, Castro, Courville, and Bellemare]{agarwal2021deep}
Rishabh Agarwal, Max Schwarzer, Pablo~Samuel Castro, Aaron Courville, and Marc~G Bellemare.
\newblock Deep reinforcement learning at the edge of the statistical precipice.
\newblock \emph{Advances in Neural Information Processing Systems}, 2021.

\bibitem[Bamford(2021)]{bamford2021griddly}
Christopher Bamford.
\newblock Griddly: A platform for ai research in games.
\newblock \emph{Software Impacts}, 8:\penalty0 100066, 2021.

\bibitem[Barto et~al.(1983)Barto, Sutton, and Anderson]{6313077}
Andrew~G. Barto, Richard~S. Sutton, and Charles~W. Anderson.
\newblock Neuronlike adaptive elements that can solve difficult learning control problems.
\newblock \emph{IEEE Transactions on Systems, Man, and Cybernetics}, SMC-13\penalty0 (5):\penalty0 834--846, 1983.
\newblock \doi{10.1109/TSMC.1983.6313077}.

\bibitem[Bellemare et~al.(2016{\natexlab{a}})Bellemare, Srinivasan, Ostrovski, Schaul, Saxton, and Munos]{bellemare2016unifying}
Marc Bellemare, Sriram Srinivasan, Georg Ostrovski, Tom Schaul, David Saxton, and Remi Munos.
\newblock Unifying count-based exploration and intrinsic motivation.
\newblock \emph{Advances in neural information processing systems}, 29, 2016{\natexlab{a}}.

\bibitem[Bellemare et~al.(2013)Bellemare, Naddaf, Veness, and Bowling]{bellemare2013arcade}
Marc~G Bellemare, Yavar Naddaf, Joel Veness, and Michael Bowling.
\newblock The arcade learning environment: An evaluation platform for general agents.
\newblock \emph{Journal of Artificial Intelligence Research}, 47:\penalty0 253--279, 2013.

\bibitem[Bellemare et~al.(2016{\natexlab{b}})Bellemare, Srinivasan, Ostrovski, Schaul, Saxton, and Munos]{10.5555/3157096.3157262}
Marc~G. Bellemare, Sriram Srinivasan, Georg Ostrovski, Tom Schaul, David Saxton, and R\'{e}mi Munos.
\newblock Unifying count-based exploration and intrinsic motivation.
\newblock In \emph{Proceedings of the 30th International Conference on Neural Information Processing Systems}, NIPS'16, pp.\  1479–1487, Red Hook, NY, USA, 2016{\natexlab{b}}. Curran Associates Inc.
\newblock ISBN 9781510838819.

\bibitem[Bengio et~al.(2009)Bengio, Louradour, Collobert, and Weston]{10.1145/1553374.1553380}
Yoshua Bengio, J\'{e}r\^{o}me Louradour, Ronan Collobert, and Jason Weston.
\newblock Curriculum learning.
\newblock In \emph{Proceedings of the 26th Annual International Conference on Machine Learning}, ICML '09, pp.\  41–48, New York, NY, USA, 2009. Association for Computing Machinery.
\newblock ISBN 9781605585161.
\newblock \doi{10.1145/1553374.1553380}.
\newblock URL \url{https://doi.org/10.1145/1553374.1553380}.

\bibitem[Berner et~al.(2019)Berner, Brockman, Chan, Cheung, D{\k{e}}biak, Dennison, Farhi, Fischer, Hashme, Hesse, et~al.]{berner2019dota}
Christopher Berner, Greg Brockman, Brooke Chan, Vicki Cheung, Przemys{\l}aw D{\k{e}}biak, Christy Dennison, David Farhi, Quirin Fischer, Shariq Hashme, Chris Hesse, et~al.
\newblock Dota 2 with large scale deep reinforcement learning.
\newblock \emph{arXiv preprint arXiv:1912.06680}, 2019.

\bibitem[Beukman et~al.(2024)Beukman, Coward, Matthews, Fellows, Jiang, Dennis, and Foerster]{beukman2024refining}
Michael Beukman, Samuel Coward, Michael Matthews, Mattie Fellows, Minqi Jiang, Michael Dennis, and Jakob Foerster.
\newblock Refining minimax regret for unsupervised environment design.
\newblock \emph{arXiv preprint arXiv:2402.12284}, 2024.

\bibitem[Bradbury et~al.(2018)Bradbury, Frostig, Hawkins, Johnson, Leary, Maclaurin, Necula, Paszke, Vander{P}las, Wanderman-{M}ilne, and Zhang]{jax2018github}
James Bradbury, Roy Frostig, Peter Hawkins, Matthew~James Johnson, Chris Leary, Dougal Maclaurin, George Necula, Adam Paszke, Jake Vander{P}las, Skye Wanderman-{M}ilne, and Qiao Zhang.
\newblock {JAX}: composable transformations of {P}ython+{N}um{P}y programs, 2018.
\newblock URL \url{http://github.com/google/jax}.

\bibitem[Brockman et~al.(2016)Brockman, Cheung, Pettersson, Schneider, Schulman, Tang, and Zaremba]{brockman2016openai}
Greg Brockman, Vicki Cheung, Ludwig Pettersson, Jonas Schneider, John Schulman, Jie Tang, and Wojciech Zaremba.
\newblock Openai gym.
\newblock \emph{arXiv preprint arXiv:1606.01540}, 2016.

\bibitem[Brown(1951)]{brown:fp1951}
George~W. Brown.
\newblock Iterative solution of games by fictitious play.
\newblock In T.~C. Koopmans (ed.), \emph{Activity Analysis of Production and Allocation}. Wiley, New York, 1951.

\bibitem[Castro(2024)]{Castro_ALE}
Pablo~Samuel Castro.
\newblock In defense of atari: The ale as a benchmark for autorl.
\newblock 2024.
\newblock URL \url{https://icml.cc/virtual/2024/39301}.

\bibitem[Chaiklin(2003)]{Chaiklin_2003}
Seth Chaiklin.
\newblock \emph{The Zone of Proximal Development in Vygotsky’s Analysis of Learning and Instruction}, pp.\  39–64.
\newblock Learning in Doing: Social, Cognitive and Computational Perspectives. Cambridge University Press, 2003.

\bibitem[Clevert et~al.(2015)Clevert, Unterthiner, and Hochreiter]{clevert2015fast}
Djork-Arn{\'e} Clevert, Thomas Unterthiner, and Sepp Hochreiter.
\newblock Fast and accurate deep network learning by exponential linear units (elus).
\newblock \emph{arXiv preprint arXiv:1511.07289}, 2015.

\bibitem[Cobbe et~al.(2020{\natexlab{a}})Cobbe, Hesse, Hilton, and Schulman]{cobbe2020leveraging}
Karl Cobbe, Chris Hesse, Jacob Hilton, and John Schulman.
\newblock Leveraging procedural generation to benchmark reinforcement learning.
\newblock In \emph{International conference on machine learning}, pp.\  2048--2056. PMLR, 2020{\natexlab{a}}.

\bibitem[Cobbe et~al.(2020{\natexlab{b}})Cobbe, Hesse, Hilton, and Schulman]{pmlr-v119-cobbe20a}
Karl Cobbe, Chris Hesse, Jacob Hilton, and John Schulman.
\newblock Leveraging procedural generation to benchmark reinforcement learning.
\newblock In Hal~Daumé III and Aarti Singh (eds.), \emph{Proceedings of the 37th International Conference on Machine Learning}, volume 119 of \emph{Proceedings of Machine Learning Research}, pp.\  2048--2056. PMLR, 13--18 Jul 2020{\natexlab{b}}.
\newblock URL \url{https://proceedings.mlr.press/v119/cobbe20a.html}.

\bibitem[Cobbe et~al.(2021)Cobbe, Hilton, Klimov, and Schulman]{cobbe2021phasic}
Karl~W Cobbe, Jacob Hilton, Oleg Klimov, and John Schulman.
\newblock Phasic policy gradient.
\newblock In \emph{International Conference on Machine Learning}, pp.\  2020--2027. PMLR, 2021.

\bibitem[Colas et~al.(2023)Colas, Teodorescu, Oudeyer, Yuan, and C{\^o}t{\'e}]{colas2023augmenting}
C{\'e}dric Colas, Laetitia Teodorescu, Pierre-Yves Oudeyer, Xingdi Yuan, and Marc-Alexandre C{\^o}t{\'e}.
\newblock Augmenting autotelic agents with large language models.
\newblock In \emph{Conference on Lifelong Learning Agents}, pp.\  205--226. PMLR, 2023.

\bibitem[Coward et~al.(2024)Coward, Beukman, and Foerster]{coward2024jaxued}
Samuel Coward, Michael Beukman, and Jakob Foerster.
\newblock Jaxued: A simple and useable ued library in jax.
\newblock \emph{arXiv preprint arXiv:2403.13091}, 2024.

\bibitem[Dennis et~al.(2020)Dennis, Jaques, Vinitsky, Bayen, Russell, Critch, and Levine]{dennis2020emergent}
Michael Dennis, Natasha Jaques, Eugene Vinitsky, Alexandre Bayen, Stuart Russell, Andrew Critch, and Sergey Levine.
\newblock Emergent complexity and zero-shot transfer via unsupervised environment design.
\newblock \emph{Advances in neural information processing systems}, 33:\penalty0 13049--13061, 2020.

\bibitem[Dharna et~al.(2022)Dharna, Summers, Dasari, Togelius, and Hoover]{dharna2022watts}
Aaron Dharna, Charlie Summers, Rohin Dasari, Julian Togelius, and Amy~K Hoover.
\newblock Watts: Infrastructure for open-ended learning.
\newblock In \emph{ICLR Workshop on Agent Learning in Open-Endedness}, 2022.

\bibitem[Du et~al.(2023{\natexlab{a}})Du, Abbeel, and Grover]{dutakes}
Yuqing Du, Pieter Abbeel, and Aditya Grover.
\newblock It takes four to tango: Multiagent self play for automatic curriculum generation.
\newblock In \emph{International Conference on Learning Representations}, 2023{\natexlab{a}}.

\bibitem[Du et~al.(2023{\natexlab{b}})Du, Watkins, Wang, Colas, Darrell, Abbeel, Gupta, and Andreas]{du2023guiding}
Yuqing Du, Olivia Watkins, Zihan Wang, C{\'e}dric Colas, Trevor Darrell, Pieter Abbeel, Abhishek Gupta, and Jacob Andreas.
\newblock Guiding pretraining in reinforcement learning with large language models.
\newblock In \emph{International Conference on Machine Learning}, pp.\  8657--8677. PMLR, 2023{\natexlab{b}}.

\bibitem[Elman(1993)]{ELMAN199371}
Jeffrey~L. Elman.
\newblock Learning and development in neural networks: the importance of starting small.
\newblock \emph{Cognition}, 48\penalty0 (1):\penalty0 71--99, 1993.
\newblock ISSN 0010-0277.
\newblock \doi{https://doi.org/10.1016/0010-0277(93)90058-4}.
\newblock URL \url{https://www.sciencedirect.com/science/article/pii/0010027793900584}.

\bibitem[Faldor et~al.(2024)Faldor, Zhang, Cully, and Clune]{faldor2024omni}
Maxence Faldor, Jenny Zhang, Antoine Cully, and Jeff Clune.
\newblock Omni-epic: Open-endedness via models of human notions of interestingness with environments programmed in code.
\newblock \emph{arXiv preprint arXiv:2405.15568}, 2024.

\bibitem[Fan et~al.(2022)Fan, Wang, Jiang, Mandlekar, Yang, Zhu, Tang, Huang, Zhu, and Anandkumar]{fan2022minedojo}
Linxi Fan, Guanzhi Wang, Yunfan Jiang, Ajay Mandlekar, Yuncong Yang, Haoyi Zhu, Andrew Tang, De-An Huang, Yuke Zhu, and Anima Anandkumar.
\newblock Minedojo: Building open-ended embodied agents with internet-scale knowledge.
\newblock \emph{Advances in Neural Information Processing Systems}, 35:\penalty0 18343--18362, 2022.

\bibitem[Florensa et~al.(2018)Florensa, Held, Geng, and Abbeel]{florensa2018automatic}
Carlos Florensa, David Held, Xinyang Geng, and Pieter Abbeel.
\newblock Automatic goal generation for reinforcement learning agents.
\newblock In \emph{International conference on machine learning}, pp.\  1515--1528. PMLR, 2018.

\bibitem[Graves et~al.(2017)Graves, Bellemare, Menick, Munos, and Kavukcuoglu]{graves2017automated}
Alex Graves, Marc~G Bellemare, Jacob Menick, Remi Munos, and Koray Kavukcuoglu.
\newblock Automated curriculum learning for neural networks.
\newblock In \emph{international conference on machine learning}, pp.\  1311--1320. Pmlr, 2017.

\bibitem[Guss et~al.(2019)Guss, Houghton, Topin, Wang, Codel, Veloso, and Salakhutdinov]{guss2019minerl}
William~H Guss, Brandon Houghton, Nicholay Topin, Phillip Wang, Cayden Codel, Manuela Veloso, and Ruslan Salakhutdinov.
\newblock Minerl: A large-scale dataset of minecraft demonstrations.
\newblock \emph{arXiv preprint arXiv:1907.13440}, 2019.

\bibitem[Haarnoja et~al.(2018)Haarnoja, Zhou, Abbeel, and Levine]{haarnoja2018soft}
Tuomas Haarnoja, Aurick Zhou, Pieter Abbeel, and Sergey Levine.
\newblock Soft actor-critic: Off-policy maximum entropy deep reinforcement learning with a stochastic actor.
\newblock In \emph{International conference on machine learning}, pp.\  1861--1870. Pmlr, 2018.

\bibitem[Hafner(2021)]{hafner2021benchmarking}
Danijar Hafner.
\newblock Benchmarking the spectrum of agent capabilities.
\newblock \emph{arXiv preprint arXiv:2109.06780}, 2021.

\bibitem[Hambro et~al.(2022{\natexlab{a}})Hambro, Mohanty, Babaev, Byeon, Chakraborty, Grefenstette, Jiang, Daejin, Kanervisto, Kim, et~al.]{hambro2022insights}
Eric Hambro, Sharada Mohanty, Dmitrii Babaev, Minwoo Byeon, Dipam Chakraborty, Edward Grefenstette, Minqi Jiang, Jo~Daejin, Anssi Kanervisto, Jongmin Kim, et~al.
\newblock Insights from the neurips 2021 nethack challenge.
\newblock In \emph{NeurIPS 2021 Competitions and Demonstrations Track}, pp.\  41--52. PMLR, 2022{\natexlab{a}}.

\bibitem[Hambro et~al.(2022{\natexlab{b}})Hambro, Raileanu, Rothermel, Mella, Rockt{\"a}schel, K{\"u}ttler, and Murray]{hambro2022dungeons}
Eric Hambro, Roberta Raileanu, Danielle Rothermel, Vegard Mella, Tim Rockt{\"a}schel, Heinrich K{\"u}ttler, and Naila Murray.
\newblock Dungeons and data: A large-scale nethack dataset.
\newblock \emph{Advances in Neural Information Processing Systems}, 35:\penalty0 24864--24878, 2022{\natexlab{b}}.

\bibitem[Heinrich \& Silver(2016)Heinrich and Silver]{heinrich2016deep}
Johannes Heinrich and David Silver.
\newblock Deep reinforcement learning from self-play in imperfect-information games.
\newblock \emph{arXiv preprint arXiv:1603.01121}, 2016.

\bibitem[Heinrich et~al.(2015{\natexlab{a}})Heinrich, Lanctot, and Silver]{10.5555/3045118.3045205}
Johannes Heinrich, Marc Lanctot, and David Silver.
\newblock Fictitious self-play in extensive-form games.
\newblock In \emph{Proceedings of the 32nd International Conference on International Conference on Machine Learning - Volume 37}, ICML'15, pp.\  805–813. JMLR.org, 2015{\natexlab{a}}.

\bibitem[Heinrich et~al.(2015{\natexlab{b}})Heinrich, Lanctot, and Silver]{Heinrich2015FictitiousSI}
Johannes Heinrich, Marc Lanctot, and David Silver.
\newblock Fictitious self-play in extensive-form games.
\newblock In \emph{International Conference on Machine Learning}, 2015{\natexlab{b}}.
\newblock URL \url{https://api.semanticscholar.org/CorpusID:13937012}.

\bibitem[Henaff et~al.(2022)Henaff, Raileanu, Jiang, and Rockt{\"a}schel]{henaff2022exploration}
Mikael Henaff, Roberta Raileanu, Minqi Jiang, and Tim Rockt{\"a}schel.
\newblock Exploration via elliptical episodic bonuses.
\newblock \emph{Advances in Neural Information Processing Systems}, 35:\penalty0 37631--37646, 2022.

\bibitem[Huang et~al.(2022)Huang, Dossa, Ye, Braga, Chakraborty, Mehta, and Ara{\'u}jo]{huang2022cleanrl}
Shengyi Huang, Rousslan Fernand~JulienDossa Dossa, Chang Ye, Jeff Braga, Dipam Chakraborty, Kinal Mehta, and Jo{\~a}o~GM Ara{\'u}jo.
\newblock Cleanrl: High-quality single-file implementations of deep reinforcement learning algorithms.
\newblock \emph{The Journal of Machine Learning Research}, 23\penalty0 (1):\penalty0 12585--12602, 2022.

\bibitem[Huang et~al.(2024)Huang, Gallou{\'e}dec, Felten, Raffin, Dossa, Zhao, Sullivan, Makoviychuk, Makoviichuk, Danesh, et~al.]{huang2024open}
Shengyi Huang, Quentin Gallou{\'e}dec, Florian Felten, Antonin Raffin, Rousslan Fernand~Julien Dossa, Yanxiao Zhao, Ryan Sullivan, Viktor Makoviychuk, Denys Makoviichuk, Mohamad~H Danesh, et~al.
\newblock Open rl benchmark: Comprehensive tracked experiments for reinforcement learning.
\newblock \emph{arXiv preprint arXiv:2402.03046}, 2024.

\bibitem[Jackson et~al.(2024)Jackson, Jiang, Parker-Holder, Vuorio, Lu, Farquhar, Whiteson, and Foerster]{jackson2024discovering}
Matthew~T Jackson, Minqi Jiang, Jack Parker-Holder, Risto Vuorio, Chris Lu, Greg Farquhar, Shimon Whiteson, and Jakob Foerster.
\newblock Discovering general reinforcement learning algorithms with adversarial environment design.
\newblock \emph{Advances in Neural Information Processing Systems}, 36, 2024.

\bibitem[Jiang et~al.(2021{\natexlab{a}})Jiang, Dennis, Parker-Holder, Foerster, Grefenstette, and Rockt{\"a}schel]{jiang2021replay}
Minqi Jiang, Michael Dennis, Jack Parker-Holder, Jakob Foerster, Edward Grefenstette, and Tim Rockt{\"a}schel.
\newblock Replay-guided adversarial environment design.
\newblock \emph{Advances in Neural Information Processing Systems}, 34:\penalty0 1884--1897, 2021{\natexlab{a}}.

\bibitem[Jiang et~al.(2021{\natexlab{b}})Jiang, Grefenstette, and Rockt{\"a}schel]{jiang2021prioritized}
Minqi Jiang, Edward Grefenstette, and Tim Rockt{\"a}schel.
\newblock Prioritized level replay.
\newblock In \emph{International Conference on Machine Learning}, pp.\  4940--4950. PMLR, 2021{\natexlab{b}}.

\bibitem[Jiang et~al.(2022)Jiang, Mediratta, Samvelyan, and Teoh]{dcd}
Minqi Jiang, Ishita Mediratta, Mikayel Samvelyan, and Jayden Teoh.
\newblock Dual curriculum design.
\newblock \url{https://github.com/facebookresearch/dcd}, 2022.

\bibitem[Jiang et~al.(2023)Jiang, Dennis, Grefenstette, and Rockt{\"a}schel]{jiang2023minimax}
Minqi Jiang, Michael Dennis, Edward Grefenstette, and Tim Rockt{\"a}schel.
\newblock minimax: Efficient baselines for autocurricula in jax.
\newblock \emph{arXiv preprint arXiv:2311.12716}, 2023.

\bibitem[Kanitscheider et~al.(2021)Kanitscheider, Huizinga, Farhi, Guss, Houghton, Sampedro, Zhokhov, Baker, Ecoffet, Tang, et~al.]{kanitscheider2021multi}
Ingmar Kanitscheider, Joost Huizinga, David Farhi, William~Hebgen Guss, Brandon Houghton, Raul Sampedro, Peter Zhokhov, Bowen Baker, Adrien Ecoffet, Jie Tang, et~al.
\newblock Multi-task curriculum learning in a complex, visual, hard-exploration domain: Minecraft.
\newblock \emph{arXiv preprint arXiv:2106.14876}, 2021.

\bibitem[Kirkpatrick et~al.(1983)Kirkpatrick, Gelatt, and Vecchi]{doi:10.1126/science.220.4598.671}
S.~Kirkpatrick, C.~D. Gelatt, and M.~P. Vecchi.
\newblock Optimization by simulated annealing.
\newblock \emph{Science}, 220\penalty0 (4598):\penalty0 671--680, 1983.
\newblock \doi{10.1126/science.220.4598.671}.
\newblock URL \url{https://www.science.org/doi/abs/10.1126/science.220.4598.671}.

\bibitem[Klink et~al.(2022)Klink, Yang, D'Eramo, Peters, and Pajarinen]{pmlr-v162-klink22a}
Pascal Klink, Haoyi Yang, Carlo D'Eramo, Jan Peters, and Joni Pajarinen.
\newblock Curriculum reinforcement learning via constrained optimal transport.
\newblock In Kamalika Chaudhuri, Stefanie Jegelka, Le~Song, Csaba Szepesvari, Gang Niu, and Sivan Sabato (eds.), \emph{Proceedings of the 39th International Conference on Machine Learning}, volume 162 of \emph{Proceedings of Machine Learning Research}, pp.\  11341--11358. PMLR, 17--23 Jul 2022.
\newblock URL \url{https://proceedings.mlr.press/v162/klink22a.html}.

\bibitem[K{\"u}ttler et~al.(2019)K{\"u}ttler, Nardelli, Lavril, Selvatici, Sivakumar, Rockt{\"a}schel, and Grefenstette]{kuttler2019torchbeast}
Heinrich K{\"u}ttler, Nantas Nardelli, Thibaut Lavril, Marco Selvatici, Viswanath Sivakumar, Tim Rockt{\"a}schel, and Edward Grefenstette.
\newblock Torchbeast: A pytorch platform for distributed rl.
\newblock \emph{arXiv preprint arXiv:1910.03552}, 2019.

\bibitem[K{\"u}ttler et~al.(2020)K{\"u}ttler, Nardelli, Miller, Raileanu, Selvatici, Grefenstette, and Rockt{\"a}schel]{kuttler2020nethack}
Heinrich K{\"u}ttler, Nantas Nardelli, Alexander Miller, Roberta Raileanu, Marco Selvatici, Edward Grefenstette, and Tim Rockt{\"a}schel.
\newblock The nethack learning environment.
\newblock \emph{Advances in Neural Information Processing Systems}, 33:\penalty0 7671--7684, 2020.

\bibitem[Lanctot et~al.(2017)Lanctot, Zambaldi, Gruslys, Lazaridou, Tuyls, P{\'e}rolat, Silver, and Graepel]{lanctot2017unified}
Marc Lanctot, Vinicius Zambaldi, Audrunas Gruslys, Angeliki Lazaridou, Karl Tuyls, Julien P{\'e}rolat, David Silver, and Thore Graepel.
\newblock A unified game-theoretic approach to multiagent reinforcement learning.
\newblock \emph{Advances in neural information processing systems}, 30, 2017.

\bibitem[Leibo et~al.(2019)Leibo, Hughes, Lanctot, and Graepel]{leibo2019autocurricula}
Joel~Z Leibo, Edward Hughes, Marc Lanctot, and Thore Graepel.
\newblock Autocurricula and the emergence of innovation from social interaction: A manifesto for multi-agent intelligence research.
\newblock \emph{arXiv preprint arXiv:1903.00742}, 2019.

\bibitem[Liang et~al.(2018)Liang, Liaw, Nishihara, Moritz, Fox, Goldberg, Gonzalez, Jordan, and Stoica]{liang2018rllib}
Eric Liang, Richard Liaw, Robert Nishihara, Philipp Moritz, Roy Fox, Ken Goldberg, Joseph Gonzalez, Michael Jordan, and Ion Stoica.
\newblock Rllib: Abstractions for distributed reinforcement learning.
\newblock In \emph{International conference on machine learning}, pp.\  3053--3062. PMLR, 2018.

\bibitem[Mediratta et~al.(2023{\natexlab{a}})Mediratta, Jiang, Parker-Holder, Dennis, Vinitsky, and Rockt{\"a}schel]{mediratta2023stabilizing}
Ishita Mediratta, Minqi Jiang, Jack Parker-Holder, Michael Dennis, Eugene Vinitsky, and Tim Rockt{\"a}schel.
\newblock Stabilizing unsupervised environment design with a learned adversary.
\newblock \emph{arXiv preprint arXiv:2308.10797}, 2023{\natexlab{a}}.

\bibitem[Mediratta et~al.(2023{\natexlab{b}})Mediratta, Jiang, Parker-Holder, Dennis, Vinitsky, and Rockt\"aschel]{pmlr-v232-mediratta23a}
Ishita Mediratta, Minqi Jiang, Jack Parker-Holder, Michael Dennis, Eugene Vinitsky, and Tim Rockt\"aschel.
\newblock Stabilizing unsupervised environment design with a learned adversary.
\newblock In Sarath Chandar, Razvan Pascanu, Hanie Sedghi, and Doina Precup (eds.), \emph{Proceedings of The 2nd Conference on Lifelong Learning Agents}, volume 232 of \emph{Proceedings of Machine Learning Research}, pp.\  270--291. PMLR, 22--25 Aug 2023{\natexlab{b}}.
\newblock URL \url{https://proceedings.mlr.press/v232/mediratta23a.html}.

\bibitem[Mehta et~al.(2020)Mehta, Diaz, Golemo, Pal, and Paull]{mehta2020active}
Bhairav Mehta, Manfred Diaz, Florian Golemo, Christopher~J Pal, and Liam Paull.
\newblock Active domain randomization.
\newblock In \emph{Conference on Robot Learning}, pp.\  1162--1176. PMLR, 2020.

\bibitem[Mella et~al.(2022)Mella, Hambro, Rothermel, and K{\"{u}}ttler]{moolib2022}
Vegard Mella, Eric Hambro, Danielle Rothermel, and Heinrich K{\"{u}}ttler.
\newblock {moolib: A Platform for Distributed RL}.
\newblock 2022.
\newblock URL \url{https://github.com/facebookresearch/moolib}.

\bibitem[Moritz et~al.(2018)Moritz, Nishihara, Wang, Tumanov, Liaw, Liang, Elibol, Yang, Paul, Jordan, et~al.]{moritz2018ray}
Philipp Moritz, Robert Nishihara, Stephanie Wang, Alexey Tumanov, Richard Liaw, Eric Liang, Melih Elibol, Zongheng Yang, William Paul, Michael~I Jordan, et~al.
\newblock Ray: A distributed framework for emerging $\{$AI$\}$ applications.
\newblock In \emph{13th USENIX symposium on operating systems design and implementation (OSDI 18)}, pp.\  561--577, 2018.

\bibitem[Narvekar et~al.(2020)Narvekar, Peng, Leonetti, Sinapov, Taylor, and Stone]{narvekar2020curriculum}
Sanmit Narvekar, Bei Peng, Matteo Leonetti, Jivko Sinapov, Matthew~E Taylor, and Peter Stone.
\newblock Curriculum learning for reinforcement learning domains: A framework and survey.
\newblock \emph{The Journal of Machine Learning Research}, 21\penalty0 (1):\penalty0 7382--7431, 2020.

\bibitem[Nikulin et~al.(2023)Nikulin, Kurenkov, Zisman, Sinii, Agarkov, and Kolesnikov]{nikulin2023xlandminigrid}
Alexander Nikulin, Vladislav Kurenkov, Ilya Zisman, Viacheslav Sinii, Artem Agarkov, and Sergey Kolesnikov.
\newblock {XL}and-minigrid: Scalable meta-reinforcement learning environments in {JAX}.
\newblock In \emph{Intrinsically-Motivated and Open-Ended Learning Workshop, NeurIPS2023}, 2023.
\newblock URL \url{https://openreview.net/forum?id=xALDC4aHGz}.

\bibitem[OpenAI et~al.(2019)OpenAI, Akkaya, Andrychowicz, Chociej, Litwin, McGrew, Petron, Paino, Plappert, Powell, Ribas, et~al.]{akkaya2019solving}
OpenAI, Ilge Akkaya, Marcin Andrychowicz, Maciek Chociej, Mateusz Litwin, Bob McGrew, Arthur Petron, Alex Paino, Matthias Plappert, Glenn Powell, Raphael Ribas, et~al.
\newblock Solving rubik's cube with a robot hand.
\newblock \emph{arXiv preprint arXiv:1910.07113}, 2019.

\bibitem[OpenAI et~al.(2021)OpenAI, Plappert, Sampedro, Xu, Akkaya, Kosaraju, Welinder, D'Sa, Petron, Pinto, et~al.]{openai2021asymmetric}
OpenAI OpenAI, Matthias Plappert, Raul Sampedro, Tao Xu, Ilge Akkaya, Vineet Kosaraju, Peter Welinder, Ruben D'Sa, Arthur Petron, Henrique P d~O Pinto, et~al.
\newblock Asymmetric self-play for automatic goal discovery in robotic manipulation.
\newblock \emph{arXiv preprint arXiv:2101.04882}, 2021.

\bibitem[Ostrovski et~al.(2017)Ostrovski, Bellemare, van~den Oord, and Munos]{10.5555/3305890.3305962}
Georg Ostrovski, Marc~G. Bellemare, A\"{a}ron van~den Oord, and R\'{e}mi Munos.
\newblock Count-based exploration with neural density models.
\newblock In \emph{Proceedings of the 34th International Conference on Machine Learning - Volume 70}, ICML'17, pp.\  2721–2730. JMLR.org, 2017.

\bibitem[Parker-Holder et~al.(2022)Parker-Holder, Jiang, Dennis, Samvelyan, Foerster, Grefenstette, and Rockt{\"a}schel]{parker2022evolving}
Jack Parker-Holder, Minqi Jiang, Michael Dennis, Mikayel Samvelyan, Jakob Foerster, Edward Grefenstette, and Tim Rockt{\"a}schel.
\newblock Evolving curricula with regret-based environment design.
\newblock In \emph{International Conference on Machine Learning}, pp.\  17473--17498. PMLR, 2022.

\bibitem[Pathak et~al.(2017{\natexlab{a}})Pathak, Agrawal, Efros, and Darrell]{10.5555/3305890.3305968}
Deepak Pathak, Pulkit Agrawal, Alexei~A. Efros, and Trevor Darrell.
\newblock Curiosity-driven exploration by self-supervised prediction.
\newblock In \emph{Proceedings of the 34th International Conference on Machine Learning - Volume 70}, ICML'17, pp.\  2778–2787. JMLR.org, 2017{\natexlab{a}}.

\bibitem[Pathak et~al.(2017{\natexlab{b}})Pathak, Agrawal, Efros, and Darrell]{pathak2017curiosity}
Deepak Pathak, Pulkit Agrawal, Alexei~A Efros, and Trevor Darrell.
\newblock Curiosity-driven exploration by self-supervised prediction.
\newblock In \emph{International conference on machine learning}, pp.\  2778--2787. PMLR, 2017{\natexlab{b}}.

\bibitem[Petrenko et~al.(2020)Petrenko, Huang, Kumar, Sukhatme, and Koltun]{petrenko2020sample}
Aleksei Petrenko, Zhehui Huang, Tushar Kumar, Gaurav Sukhatme, and Vladlen Koltun.
\newblock Sample factory: Egocentric 3d control from pixels at 100000 fps with asynchronous reinforcement learning.
\newblock In \emph{International Conference on Machine Learning}, pp.\  7652--7662. PMLR, 2020.

\bibitem[Piterbarg et~al.(2024)Piterbarg, Pinto, and Fergus]{piterbarg2024nethack}
Ulyana Piterbarg, Lerrel Pinto, and Rob Fergus.
\newblock Nethack is hard to hack.
\newblock \emph{Advances in Neural Information Processing Systems}, 36, 2024.

\bibitem[Portelas et~al.(2020{\natexlab{a}})Portelas, Colas, Hofmann, and Oudeyer]{portelas2020teacher}
R{\'e}my Portelas, C{\'e}dric Colas, Katja Hofmann, and Pierre-Yves Oudeyer.
\newblock Teacher algorithms for curriculum learning of deep rl in continuously parameterized environments.
\newblock In \emph{Conference on Robot Learning}, pp.\  835--853. PMLR, 2020{\natexlab{a}}.

\bibitem[Portelas et~al.(2020{\natexlab{b}})Portelas, Colas, Weng, Hofmann, and Oudeyer]{portelas2020automatic}
R{\'e}my Portelas, C{\'e}dric Colas, Lilian Weng, Katja Hofmann, and Pierre-Yves Oudeyer.
\newblock Automatic curriculum learning for deep rl: A short survey.
\newblock \emph{arXiv preprint arXiv:2003.04664}, 2020{\natexlab{b}}.

\bibitem[Racaniere et~al.(2019)Racaniere, Lampinen, Santoro, Reichert, Firoiu, and Lillicrap]{racaniere2019automated}
Sebastien Racaniere, Andrew~K Lampinen, Adam Santoro, David~P Reichert, Vlad Firoiu, and Timothy~P Lillicrap.
\newblock Automated curricula through setter-solver interactions.
\newblock \emph{arXiv preprint arXiv:1909.12892}, 2019.

\bibitem[Raffin et~al.(2021)Raffin, Hill, Gleave, Kanervisto, Ernestus, and Dormann]{stable-baselines3}
Antonin Raffin, Ashley Hill, Adam Gleave, Anssi Kanervisto, Maximilian Ernestus, and Noah Dormann.
\newblock Stable-baselines3: Reliable reinforcement learning implementations.
\newblock \emph{Journal of Machine Learning Research}, 22\penalty0 (268):\penalty0 1--8, 2021.
\newblock URL \url{http://jmlr.org/papers/v22/20-1364.html}.

\bibitem[Rasmussen(1999)]{NIPS1999_97d98119}
Carl Rasmussen.
\newblock The infinite gaussian mixture model.
\newblock In S.~Solla, T.~Leen, and K.~M\"{u}ller (eds.), \emph{Advances in Neural Information Processing Systems}, volume~12. MIT Press, 1999.
\newblock URL \url{https://proceedings.neurips.cc/paper_files/paper/1999/file/97d98119037c5b8a9663cb21fb8ebf47-Paper.pdf}.

\bibitem[Romac et~al.(2021)Romac, Portelas, Hofmann, and Oudeyer]{romac2021teachmyagent}
Cl{\'e}ment Romac, R{\'e}my Portelas, Katja Hofmann, and Pierre-Yves Oudeyer.
\newblock Teachmyagent: a benchmark for automatic curriculum learning in deep rl.
\newblock In \emph{International Conference on Machine Learning}, pp.\  9052--9063. PMLR, 2021.

\bibitem[Rosseau et~al.(2022)Rosseau, de~Escobar, and Nowe]{rosseau2022toward}
Andries Rosseau, Raphael Avalos~Martinez de~Escobar, and Ann Nowe.
\newblock Toward evolutionary autocurricula: Emergent sociality from inclusive rewards.
\newblock In \emph{From Cells to Societies: Collective Learning across Scales}, 2022.
\newblock URL \url{https://openreview.net/forum?id=BcUNSzkT-c}.

\bibitem[Rutherford et~al.(2024)Rutherford, Beukman, Willi, Lacerda, Hawes, and Foerster]{rutherfordno}
Alexander Rutherford, Michael Beukman, Timon Willi, Bruno Lacerda, Nick Hawes, and Jakob~Nicolaus Foerster.
\newblock No regrets: Investigating and improving regret approximations for curriculum discovery.
\newblock In \emph{The Thirty-eighth Annual Conference on Neural Information Processing Systems}, 2024.

\bibitem[Salimans \& Chen(2018)Salimans and Chen]{salimans2018learning}
Tim Salimans and Richard Chen.
\newblock Learning montezuma's revenge from a single demonstration.
\newblock \emph{arXiv preprint arXiv:1812.03381}, 2018.

\bibitem[Samuel(1959)]{5392560}
A.~L. Samuel.
\newblock Some studies in machine learning using the game of checkers.
\newblock \emph{IBM Journal of Research and Development}, 3\penalty0 (3):\penalty0 210--229, 1959.
\newblock \doi{10.1147/rd.33.0210}.

\bibitem[Samvelyan et~al.(2022)Samvelyan, Khan, Dennis, Jiang, Parker-Holder, Foerster, Raileanu, and Rockt{\"a}schel]{samvelyan2022maestro}
Mikayel Samvelyan, Akbir Khan, Michael~D Dennis, Minqi Jiang, Jack Parker-Holder, Jakob~Nicolaus Foerster, Roberta Raileanu, and Tim Rockt{\"a}schel.
\newblock Maestro: Open-ended environment design for multi-agent reinforcement learning.
\newblock In \emph{The Eleventh International Conference on Learning Representations}, 2022.

\bibitem[Schaul et~al.(2011)Schaul, Togelius, and Schmidhuber]{schaul2011measuring}
Tom Schaul, Julian Togelius, and J{\"u}rgen Schmidhuber.
\newblock Measuring intelligence through games.
\newblock \emph{arXiv preprint arXiv:1109.1314}, 2011.

\bibitem[Schulman et~al.(2015)Schulman, Levine, Abbeel, Jordan, and Moritz]{pmlr-v37-schulman15}
John Schulman, Sergey Levine, Pieter Abbeel, Michael Jordan, and Philipp Moritz.
\newblock Trust region policy optimization.
\newblock In Francis Bach and David Blei (eds.), \emph{Proceedings of the 32nd International Conference on Machine Learning}, volume~37 of \emph{Proceedings of Machine Learning Research}, pp.\  1889--1897, Lille, France, 07--09 Jul 2015. PMLR.
\newblock URL \url{https://proceedings.mlr.press/v37/schulman15.html}.

\bibitem[Schulman et~al.(2017)Schulman, Wolski, Dhariwal, Radford, and Klimov]{schulman2017proximal}
John Schulman, Filip Wolski, Prafulla Dhariwal, Alec Radford, and Oleg Klimov.
\newblock Proximal policy optimization algorithms.
\newblock \emph{arXiv preprint arXiv:1707.06347}, 2017.

\bibitem[Silver et~al.(2016)Silver, Huang, Maddison, Guez, Sifre, Driessche, Schrittwieser, Antonoglou, Panneershelvam, Lanctot, Dieleman, Grewe, Nham, Kalchbrenner, Sutskever, Lillicrap, Leach, Kavukcuoglu, Graepel, and Hassabis]{alphago}
David Silver, Aja Huang, Christopher Maddison, Arthur Guez, Laurent Sifre, George Driessche, Julian Schrittwieser, Ioannis Antonoglou, Veda Panneershelvam, Marc Lanctot, Sander Dieleman, Dominik Grewe, John Nham, Nal Kalchbrenner, Ilya Sutskever, Timothy Lillicrap, Madeleine Leach, Koray Kavukcuoglu, Thore Graepel, and Demis Hassabis.
\newblock Mastering the game of go with deep neural networks and tree search.
\newblock \emph{Nature}, 529:\penalty0 484--489, 01 2016.
\newblock \doi{10.1038/nature16961}.

\bibitem[Silver et~al.(2017)Silver, Hubert, Schrittwieser, Antonoglou, Lai, Guez, Lanctot, Sifre, Kumaran, Graepel, et~al.]{silver2017mastering}
David Silver, Thomas Hubert, Julian Schrittwieser, Ioannis Antonoglou, Matthew Lai, Arthur Guez, Marc Lanctot, Laurent Sifre, Dharshan Kumaran, Thore Graepel, et~al.
\newblock Mastering chess and shogi by self-play with a general reinforcement learning algorithm.
\newblock \emph{arXiv preprint arXiv:1712.01815}, 2017.

\bibitem[Suarez(2023)]{Suarez23:PufferLib}
Joseph Suarez.
\newblock {P}uffer{L}ib: Making reinforcement learning libraries and environments play nice.
\newblock In \emph{Agent Learning in Open-Endedness Workshop at NeurIPS '23}, 2023.

\bibitem[Suarez et~al.(2019)Suarez, Du, Isola, and Mordatch]{suarez2019neural}
Joseph Suarez, Yilun Du, Phillip Isola, and Igor Mordatch.
\newblock Neural mmo: A massively multiagent game environment for training and evaluating intelligent agents.
\newblock \emph{arXiv preprint arXiv:1903.00784}, 2019.

\bibitem[Suarez et~al.(2024)Suarez, Bloomin, Choe, Li, Sullivan, Kanna, Scott, Shuman, Bradley, Castricato, et~al.]{suarez2024neural}
Joseph Suarez, David Bloomin, Kyoung~Whan Choe, Hao~Xiang Li, Ryan Sullivan, Nishaanth Kanna, Daniel Scott, Rose Shuman, Herbie Bradley, Louis Castricato, et~al.
\newblock Neural mmo 2.0: A massively multi-task addition to massively multi-agent learning.
\newblock \emph{Advances in Neural Information Processing Systems}, 36, 2024.

\bibitem[Sukhbaatar et~al.(2018)Sukhbaatar, Lin, Kostrikov, Synnaeve, Szlam, and Fergus]{sukhbaatar2018intrinsic}
Sainbayar Sukhbaatar, Zeming Lin, Ilya Kostrikov, Gabriel Synnaeve, Arthur Szlam, and Rob Fergus.
\newblock Intrinsic motivation and automatic curricula via asymmetric self-play.
\newblock In \emph{6th International Conference on Learning Representations, ICLR 2018}, 2018.

\bibitem[Taiga et~al.(2021)Taiga, Fedus, Machado, Courville, and Bellemare]{taiga2021bonus}
Adrien~Ali Taiga, William Fedus, Marlos~C Machado, Aaron Courville, and Marc~G Bellemare.
\newblock On bonus-based exploration methods in the arcade learning environment.
\newblock \emph{arXiv preprint arXiv:2109.11052}, 2021.

\bibitem[Taylor \& Stone(2009)Taylor and Stone]{10.5555/1577069.1755839}
Matthew~E. Taylor and Peter Stone.
\newblock Transfer learning for reinforcement learning domains: A survey.
\newblock \emph{J. Mach. Learn. Res.}, 10:\penalty0 1633–1685, December 2009.
\newblock ISSN 1532-4435.

\bibitem[Team et~al.(2023)Team, Bauer, Baumli, Baveja, Behbahani, Bhoopchand, Bradley-Schmieg, Chang, Clay, Collister, et~al.]{team2023human}
Adaptive~Agent Team, Jakob Bauer, Kate Baumli, Satinder Baveja, Feryal Behbahani, Avishkar Bhoopchand, Nathalie Bradley-Schmieg, Michael Chang, Natalie Clay, Adrian Collister, et~al.
\newblock Human-timescale adaptation in an open-ended task space.
\newblock \emph{arXiv preprint arXiv:2301.07608}, 2023.

\bibitem[Terry et~al.(2021)Terry, Black, Grammel, Jayakumar, Hari, Sullivan, Santos, Dieffendahl, Horsch, Perez-Vicente, et~al.]{terry2021pettingzoo}
JK~Terry, Benjamin~J Black, Nathaniel Grammel, Mario Jayakumar, Ananth Hari, Ryan Sullivan, Luis Santos, Clemens Dieffendahl, Caroline Horsch, Rodrigo De~Lazcano Perez-Vicente, et~al.
\newblock Pettingzoo: Gym for multi-agent reinforcement learning.
\newblock In \emph{Advances in Neural Information Processing Systems}, 2021.

\bibitem[Tesauro(1995)]{10.1145/203330.203343}
Gerald Tesauro.
\newblock Temporal difference learning and td-gammon.
\newblock \emph{Commun. ACM}, 38\penalty0 (3):\penalty0 58–68, March 1995.
\newblock ISSN 0001-0782.
\newblock \doi{10.1145/203330.203343}.
\newblock URL \url{https://doi.org/10.1145/203330.203343}.

\bibitem[Tobin et~al.(2017)Tobin, Fong, Ray, Schneider, Zaremba, and Abbeel]{tobin2017domain}
Josh Tobin, Rachel Fong, Alex Ray, Jonas Schneider, Wojciech Zaremba, and Pieter Abbeel.
\newblock Domain randomization for transferring deep neural networks from simulation to the real world.
\newblock In \emph{2017 IEEE/RSJ international conference on intelligent robots and systems (IROS)}, pp.\  23--30. IEEE, 2017.

\bibitem[Towers et~al.(2024)Towers, Kwiatkowski, Terry, Balis, De~Cola, Deleu, Goulao, Kallinteris, Krimmel, KG, et~al.]{towers2024gymnasium}
Mark Towers, Ariel Kwiatkowski, Jordan Terry, John~U Balis, Gianluca De~Cola, Tristan Deleu, Manuel Goulao, Andreas Kallinteris, Markus Krimmel, Arjun KG, et~al.
\newblock Gymnasium: A standard interface for reinforcement learning environments.
\newblock \emph{arXiv preprint arXiv:2407.17032}, 2024.

\bibitem[Tuyls et~al.(2023)Tuyls, Madeka, Torkkola, Foster, Narasimhan, and Kakade]{tuyls2023scaling}
Jens Tuyls, Dhruv Madeka, Kari Torkkola, Dean Foster, Karthik Narasimhan, and Sham Kakade.
\newblock Scaling laws for imitation learning in nethack.
\newblock \emph{arXiv preprint arXiv:2307.09423}, 2023.

\bibitem[Tzannetos et~al.(2023)Tzannetos, Gomes~Ribeiro, Kamalaruban, and Singla]{tzannetos2023proximal}
George Tzannetos, B{\'a}rbara Gomes~Ribeiro, Parameswaran Kamalaruban, and Adish Singla.
\newblock Proximal curriculum for reinforcement learning agents.
\newblock \emph{Transactions on Machine Learning Research}, 2023\penalty0 (5):\penalty0 1--21, 2023.

\bibitem[Vinyals et~al.(2019)Vinyals, Babuschkin, Czarnecki, Mathieu, Dudzik, Chung, Choi, Powell, Ewalds, Georgiev, Oh, Horgan, Kroiss, Danihelka, Huang, Sifre, Cai, Agapiou, Jaderberg, Vezhnevets, Leblond, Pohlen, Dalibard, Budden, Sulsky, Molloy, Paine, Gulcehre, Wang, Pfaff, Wu, Ring, Yogatama, W{\"u}nsch, McKinney, Smith, Schaul, Lillicrap, Kavukcuoglu, Hassabis, Apps, and Silver]{Vinyals2019GrandmasterLI}
Oriol Vinyals, Igor Babuschkin, Wojciech~M. Czarnecki, Micha{\"e}l Mathieu, Andrew Dudzik, Junyoung Chung, David~H. Choi, Richard Powell, Timo Ewalds, Petko Georgiev, Junhyuk Oh, Dan Horgan, Manuel Kroiss, Ivo Danihelka, Aja Huang, L.~Sifre, Trevor Cai, John~P. Agapiou, Max Jaderberg, Alexander~Sasha Vezhnevets, R{\'e}mi Leblond, Tobias Pohlen, Valentin Dalibard, David Budden, Yury Sulsky, James Molloy, Tom~Le Paine, Caglar Gulcehre, Ziyun Wang, Tobias Pfaff, Yuhuai Wu, Roman Ring, Dani Yogatama, Dario W{\"u}nsch, Katrina McKinney, Oliver Smith, Tom Schaul, Timothy~P. Lillicrap, Koray Kavukcuoglu, Demis Hassabis, Chris Apps, and David Silver.
\newblock Grandmaster level in starcraft ii using multi-agent reinforcement learning.
\newblock \emph{Nature}, 575:\penalty0 350 -- 354, 2019.
\newblock URL \url{https://api.semanticscholar.org/CorpusID:204972004}.

\bibitem[Vygotsky(1978)]{ef4d7fb0-848f-3480-8634-d49a5f5c57df}
L.~S. Vygotsky.
\newblock \emph{Mind in Society: Development of Higher Psychological Processes}.
\newblock Harvard University Press, 1978.
\newblock ISBN 9780674576285.
\newblock URL \url{http://www.jstor.org/stable/j.ctvjf9vz4}.

\bibitem[Wang et~al.(2019)Wang, Lehman, Clune, and Stanley]{wang2019paired}
Rui Wang, Joel Lehman, Jeff Clune, and Kenneth~O Stanley.
\newblock Paired open-ended trailblazer (poet): Endlessly generating increasingly complex and diverse learning environments and their solutions.
\newblock \emph{arXiv preprint arXiv:1901.01753}, 2019.

\bibitem[Wang et~al.(2020)Wang, Lehman, Rawal, Zhi, Li, Clune, and Stanley]{wang2020enhanced}
Rui Wang, Joel Lehman, Aditya Rawal, Jiale Zhi, Yulun Li, Jeff Clune, and Kenneth~O Stanley.
\newblock Enhanced poet: open-ended reinforcement learning through unbounded invention of learning challenges and their solutions.
\newblock In \emph{Proceedings of the 37th International Conference on Machine Learning}, pp.\  9940--9951, 2020.

\bibitem[Wurman et~al.(2022)Wurman, Barrett, Kawamoto, MacGlashan, Subramanian, Walsh, Capobianco, Devlic, Eckert, Fuchs, Gilpin, Khandelwal, Kompella, Lin, MacAlpine, Oller, Seno, Sherstan, Thomure, and Kitano]{gtsophy}
Peter Wurman, Samuel Barrett, Kenta Kawamoto, James MacGlashan, Kaushik Subramanian, Thomas Walsh, Roberto Capobianco, Alisa Devlic, Franziska Eckert, Florian Fuchs, Leilani Gilpin, Piyush Khandelwal, Varun Kompella, HaoChih Lin, Patrick MacAlpine, Declan Oller, Takuma Seno, Craig Sherstan, Michael Thomure, and Hiroaki Kitano.
\newblock Outracing champion gran turismo drivers with deep reinforcement learning.
\newblock \emph{Nature}, 602:\penalty0 223--228, 02 2022.
\newblock \doi{10.1038/s41586-021-04357-7}.

\bibitem[Yuan et~al.(2024)Yuan, Castanyer, Li, Jin, Berseth, and Zeng]{yuan_roger2024rlexplore}
Mingqi Yuan, Roger~Creus Castanyer, Bo~Li, Xin Jin, Glen Berseth, and Wenjun Zeng.
\newblock Rlexplore: Accelerating research in intrinsically-motivated reinforcement learning.
\newblock \emph{arXiv preprint arXiv:2405.19548}, 2024.

\bibitem[Zhang et~al.(2023)Zhang, Lehman, Stanley, and Clune]{zhang2023omni}
Jenny Zhang, Joel Lehman, Kenneth Stanley, and Jeff Clune.
\newblock Omni: Open-endedness via models of human notions of interestingness.
\newblock \emph{arXiv preprint arXiv:2306.01711}, 2023.

\bibitem[Zhang et~al.(2024)Zhang, Xu, Ma, Yu, Tu, Huang, Ye, Ding, Yang, and Wang]{zhang2024survey}
Ruize Zhang, Zelai Xu, Chengdong Ma, Chao Yu, Wei-Wei Tu, Shiyu Huang, Deheng Ye, Wenbo Ding, Yaodong Yang, and Yu~Wang.
\newblock A survey on self-play methods in reinforcement learning.
\newblock \emph{arXiv preprint arXiv:2408.01072}, 2024.

\bibitem[Zhu et~al.(2023)Zhu, Lin, Jain, and Zhou]{zhu2023transfer}
Zhuangdi Zhu, Kaixiang Lin, Anil~K Jain, and Jiayu Zhou.
\newblock Transfer learning in deep reinforcement learning: A survey.
\newblock \emph{IEEE Transactions on Pattern Analysis and Machine Intelligence}, 2023.

\end{thebibliography}
\bibliographystyle{rlj}

\appendix
\beginSupplementaryMaterials






\section{API Reference}

\subsection{Curriculum API}
\label{app:curriculum_api}

\begin{figure}[!ht]
\label{listing:curriculum}
\begin{minted}[frame=single, linenos, fontfamily=cmss]{python}
class Curriculum:
    """API for defining curricula to interface with Gym environments."""

    def _sample_distribution(self) -> List[float]:
        """ Returns a sample distribution over the task space.
            Any curriculum that maintains a true probability distribution
            should implement this method to retrieve the distribution. """

    def sample(self, k: int = 1) -> List[Any]:
        """ Sample k tasks from the curriculum. """

    def update_task_progress(self, task: Any, progress: Tuple[float, bool])
        """ Update the curriculum with a task and its progress. 
            Progress is defined by the environment's TaskWrapper. """

    def update_on_step(self, obs: Any, rew: float, done: bool, info: dict):
        """ Update the curriculum with the environment outputs
            for the most recent step. """

    def update_on_episode(self, return: float, length: int, task: Any, env_id: int = None):
        """Update the curriculum with episode results from the environment."""

    def update_on_demand(self, metrics: Dict):
        """ Update the curriculum with arbitrary inputs.
            Typically used to incorporate gradient or error-based
            metrics from the training process. """

    def get_opponent(self, opponent_id: int) -> Agent:
        """ Load the agent corresponding to the given opponent_id. """

    def update_agent(self, opponent: Agent):
        """ Add opponent to agent store. """

    def update_winrate(self, opponent_id: int, opponent_return: int):
        """ Update the winrate for the given opponent_id based on environment returns. """
\end{minted}
\caption{An abbreviated summary of the Curriculum interface. These represent the main methods for updating and sampling from a curriculum in Syllabus. The get\_opponent, update\_agent, and update\_winrate methods are used for self-play.}
\end{figure}
\clearpage

\subsection{Task Space API}
\label{app:taskspace_api}

\begin{figure}[!ht]
\label{listing:taskspace}
\begin{minted}[frame=single, linenos, fontfamily=cmss]{python}
class TaskSpace:
    """ API for the range of tasks supported by an environment or 
         curriculum learning algorithm."""

    def sample(self) -> Any:
        """ Sample a task randomly from the space. """

    def decode(self, encoding: Any) -> Any:
        """ Decode the task encoding to a format that can be interpreted by 
             the environment. The task will be passed to the environment in this format."""

    def encode(self, task: Any) -> Any:
        """ Convert the task to an efficient encoding that can be interpreted by 
             the curriculum. All curriculum updates receive the task in this format."""

    def seed(self, seed: int):
        """ Seed the task space for deterministic sampling. """

    @property
    def tasks(self) -> List[Any]:
       """ Returns a list of all tasks if task space is discrete """
\end{minted}
\caption{Main features of the Task Space API.}
\end{figure}

\subsection{Task Interface API}
\label{app:task_api}

In unsupervised environment design, we study underspecified POMDPs (UPOMDPs), which have free configuration variables that need to be chosen to produce a fully specified POMDP \citep{dennis2020emergent}. In multi-task environments these free variables are the task, which can be a level seed, map specification, reward function, or even a new environment instance with different dynamics. Syllabus supports UPOMDPs by adding a \texttt{new\_task} argument in the reset function of the standard Gym API. However, most environments do not support this behavior by default. We provide a TaskWrapper that accepts the new task, reconfigures the environment, then resets the environment for the next episode. This even allows us to add multi-task capabilities to single-task environments.

The Task Interface can also define an environment-specific progress metric to support curricula that depend on task success rates, and encode the current task into the observation space for task-conditional policies. As a result of this API structure, Syllabus is by far the easiest curriculum learning library to incorporate with preexisting training code, needing only a few lines of code to use complex ACL algorithms. Swapping between different curricula often requires only a single line of code change. Users can integrate new environments with a simple wrapper that tells Syllabus how to interpret its task space. We encourage the reader to look at our Procgen training script to better understand what Syllabus code looks like in practice. The next section explains the infrastructure that allows us to maintain this simplicity with asynchronous RL training infrastructure.

\begin{figure}[!ht]
\label{listing:taskwrapper}
\begin{minted}[frame=single, linenos, fontfamily=cmss]{python}
class TaskWrapper:
    """ Interface that changes the task assigned to a 
         Gym or PettingZoo environment during reset."""

        def reset(self, *args, new_task: Any = None, **kwargs):
             """ Accepts a new task to be used in the next episode. """

        def change_task(self, new_task: Any):
            """ Modify the environment to use the new task. """

        def _task_completion(self, obs, rew, term, trunc, info) -> float:
            """ Implement this function to define a task progress metric. """

        def _encode_goal(self) -> Any:
            """ Encode the goal for the agent to observe. """

\end{minted}
\caption{Main features of the Task Interface.}
\end{figure}

\subsection{Sequential Curriculum}
Syllabus implements a \texttt{Sequential} curriculum, which can be though of as a meta-curriculum over individual \texttt{Curriculum} objects. The curriculum has multiple stages, each of which will run until user-defined stopping conditions are met, after which the curriculum will move on to the next stage. The curriculum passes any updates that it receives to the current curriculum stage, so we can even use automatic curricula sequentially.

It is initialized with a set of stages, which can be individual elements of the task space, lists of tasks, entire task spaces, or other \texttt{Curriculum} objects. Individual tasks will be converted into a \texttt{Constant} curriculum which always returns the same task. Lists of tasks or task spaces are converted to \texttt{DomainRandomization} curricula over the provided tasks.

The stopping conditions can be predefined numbers of steps, tasks, or episodes, or they can be episodic return thresholds. We use a convenient text-based interface for defining stopping conditions which supports composite conditions. For example, the user can pass \texttt{"return>=1.0\&\&episodes>=1000"} as a valid composite conditions, which will stop the current stage once the agent has experiences 1000 episodes and achieves at least 1.0 return. The sequential curriculum receives updates from the environments to track agent progress automatically. 

For an example of this sequential curriculum in practice, see \autoref{app:nmmo-sequential}.

\section{Optimization}
\label{app:optimization}
As a consequence of the choice to use a separate multiprocessing system from the RL training loop, Syllabus incurs some unavoidable computational costs. Specifically, receiving and sending information in the environments decreases the effective steps per second of each environment, while sampling and sending tasks in the actor process increases the computational load on the main process. We evaluate Syllabus with NetHack to demonstrate the effect on overall steps per second. We use a minimal curriculum that always returns the same task to isolate the impact of our multiprocessing infrastructure. 

The results are shown in \autoref{tab:performance} for 128 environments each running 64 episodes on a 2.20GHz 32-core Intel i9-13950HX \footnote{Syllabus is under continuous development, so these numbers may not reflect the performance of the most recent version of the library. They are accurate as of the time of publication.}. We test Syllabus using both Python's native multiprocessing package and Ray \citep{moritz2018ray}. Syllabus skips per-step updates for curricula that do not require them, instead only sending episodic data and tasks at the end of each episode. 

Note that this is a worst case test for several reasons. Typically in asynchronous reinforcement learning, environments are vectorized and stepped together, such that N environments step at the speed of the single slowest environment. Here we run each environment independently, so they are not bottlenecked by vectorization. The NLE is an extremely fast environment with large observations, which stresses the multiprocessing communication bandwidth. Finally, RL is usually bottle-necked by policy optimization rather than environment iteration time. We expect Syllabus's impact on performance (as a percentage of total computation) to be much lower for more computationally intensive environments and when used in a real RL training context.

\begin{table}[h]
    \centering
    \caption{Syllabus Performance Costs}
    \begin{tabular}{@{}lccc@{}}
        \toprule
        Multiprocessing & Without Syllabus & Episodic Updates & Step Updates \\
        \midrule
        Native Python & 125s & 131s (+4.8\%) & 150.0s (+20\%) \\
        Ray & 135s & 150s (+11.1\%) & 161s (+19.3\%) \\
        \bottomrule

    \end{tabular} 
    \vspace{0.1cm}
    \label{tab:performance}
\end{table}

\section{Testing}
\label{app:testing}

We use pytest to continuously test and benchmark the performance of new additions to Syllabus on several environments. The multiprocessing infrastructure is evaluated to ensure that every sampled task is received by the environments and every environment updated is processed by the curricula, as well as several other safeguards. We use unit tests for task spaces and core curriculum features. We compare the performance of our algorithm implementations against the original implementations or original paper results whenever possible.

\clearpage
\section{Experiments}
\label{app:experiments}

This section outlines the details of our experimental setup for each environment. All of the code for these experiments is also open-sourced on GitHub. Reinforcement learning research typically compares training returns to evaluate agents, but this is not valid when using curriculum learning. Curriculum learning modifies the training task distribution, meaning that higher returns may indicate that the curriculum prioritized easier tasks or tasks with larger return scales. In each experiment we separately evaluate agents on uniformly sampled tasks.

The experiments in this paper were run on consumer GPUs, mainly the GTX 1080Ti, RTX 2080Ti, and GTX Titan X GPUs. Each experiment was run on a single GPU. An estimate of the total cost of experiments for each environment is listed in \autoref{tab:compute}. Note that this is only the cost to reproduce the exact plots in the main paper, not including any supplementary materials, hyperparameter tuning, or development time, which would make these estimates many times larger.

\begin{table}[h]
    \centering
    \caption{Experiment Compute Resources}
    \begin{tabular}{@{}lcccc@{}}
        \toprule
        & \multicolumn{2}{c}{Per Experiment} & \multicolumn{2}{c}{Total} \\
        \midrule
        & CPU Hours & GPU Hours & CPU Hours & GPU Hours \\
        \midrule
        Procgen & 110 & 14 & 33000 & 4200 \\
        Crafter & 224 & 14 & 6720 & 420 \\
        Neural MMO & 352 & 22 & 7040 & 440 \\
        NetHack & 960 & 60 & 19200 & 1200 \\
        LaserTag & 192 & 12 & 2880 & 180 \\
        \bottomrule
    \end{tabular} 
    \vspace{0.1cm}
    \caption{Approximate cost to reproduce experiments in main paper.}
    \label{tab:compute}
\end{table}

\subsection{Procgen}
\label{app:procgen}

Our Procgen experiments use the same ResNet architecture and hyperparameters as previous work \citep{cobbe2020leveraging, jiang2021prioritized}. Procgen uses an action space of 15 discrete actions and produces e 64 × 64 × 3 RGB observations.  We use the exact same ResBlock model architecture, PPO hyperparameters, and PLR options as \citet{cobbe2020leveraging} and \citet{jiang2021prioritized} to reproduce their results with Syllabus's implementation of PLR. We experiment on a subset of 10 Procgen environments: Bigfish, Bossfight, Caveflyer, Chaser, Climber, Dodgeball, Fruitbot, Leaper, Ninja, and Plunder. We train 5 agents per environment on 200 seeds of the easy level distribution for 25M steps and evaluate them on the full distribution of seeds for 10 episodes every 16,384 environment steps. We compute normalized returns using the maximum and minimum return values for each environment listed in \citet{cobbe2020leveraging} according to the formula $r_N = \frac{r - r_{min}}{r_{max} - r_{min}}$. This allows us to weigh each environment equally while aggregating returns, such as in \autoref{fig:procgen}. For the LP, SFL, and OMNI curricula we compute the task reward as $min(max(r_N, 0.0), 1.0)$.

\begin{figure}[h!]
  \centering
  \includegraphics[width=1.0\textwidth]{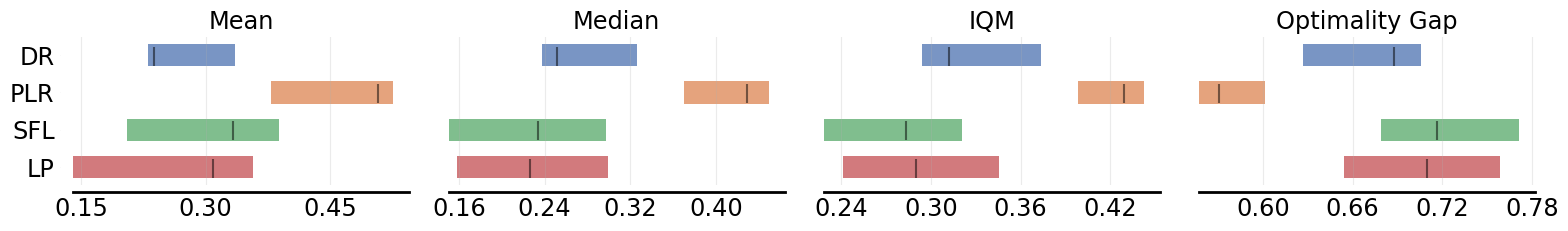}
   
  \caption{95\% Stratified Boostrapped Confidence Intervals for the Mean, Median, Interquartile Mean, and Optimality Gap of Normalized Test Returns of each ACL algorithm on Procgen.}
  \label{fig:procgen_rliable}
  \vspace{0.5cm}
\end{figure}

\clearpage

\begin{figure}[h!]
  \centering
  \includegraphics[width=1.0\textwidth]{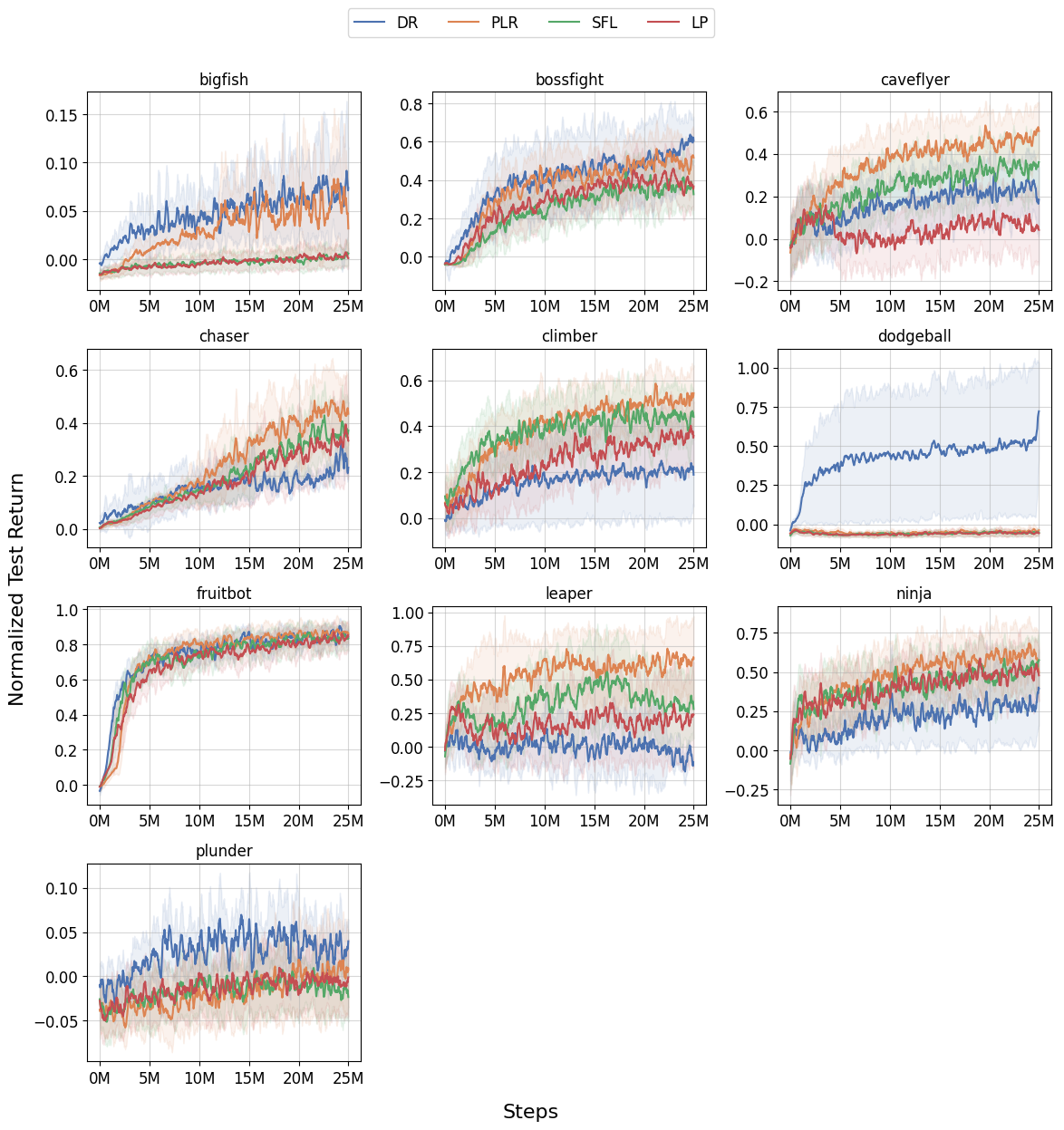}
   
  \caption{Normalized test returns for Domain Randomization, Prioritized Level Replay, Sampling for Learnability, and Learning Progress evaluated on 10 Procgen environments.}
  \label{fig:procgen_all}
  \vspace{0.5cm}
\end{figure}
\clearpage

\subsection{Crafter}
\label{app:crafter}

For our Crafter experiments, we add Syllabus to the open-source training code from \citet{zhang2023omni}, meaning we use the same model architecture, hyperparameters, and Learning Progress arguments. Crafter has 13 discrete actions and 64 x 64 x 3 RGB observations. The observations are passed through a 2-layer CNN with ReLU activations followed by a fully connected layer of size 256. These visual embeddings are concatenated with a task encoding then passed into a 256 node LSTM layer. The actions and values are generated by separate 2-layer linear heads. We train agents for 10M steps and evaluate them every 25 updates (819,200 steps) on the full task space (excluding impossible tasks, which are assigned a success rate of 0). 

As in \citet{zhang2023omni} we use a task space of 15 unique tasks \{"collect\_wood", "collect\_stone", "collect\_coal",  "collect\_iron", "collect\_diamond", "collect\_drink", "make\_iron\_pickaxe", "make\_iron\_sword", "make\_stone\_pickaxe", "make\_stone\_sword", "make\_wood\_pickaxe", "make\_wood\_sword", "place\_furnace", "place\_stone", "place\_table"\}, 90 repeat tasks, including 10 repeat tasks for each "collect" task (e.g. "collect\_9\_coal"), and 5 repeat tasks each for "make" and "place" tasks, as well as 1024 impossible tasks. The impossible tasks have a fixed success rate of 0, and serve as distractions for the curriculum. This is a realistic representation of challenging tasks where most of the task space will be inaccessible to the agent until it learns advanced skills. Each ACL algorithm quickly learns to ignore these tasks while Domain Randomization samples them with the same frequency as the 105 real tasks. In this paper we report success rates including the impossible tasks, but the success rate over possible tasks can be recovered by multiplying our reported metrics by ~10.75.

As in \citet{kanitscheider2021multi} and \citet{zhang2023omni} we train agent on the Simon Says task where the agent has 25 minutes (1500 steps) to complete as many tasks as possible in an episode. The agent has at most 5 minutes (300 steps) to complete any one task. If the agent succeeds, it is given a reward of 1 and assigned a new task. If it fails to complete the task in the time limit, it receives a reward of -1 and is assigned a new task. This Simon Says task in particular requires special consideration because we need to sample new tasks mid-episode, not just during the environment reset. Syllabus supports this behavior through it's synchronization wrappers.

\begin{figure}[h!]
  \centering
  \includegraphics[width=1.0\textwidth]{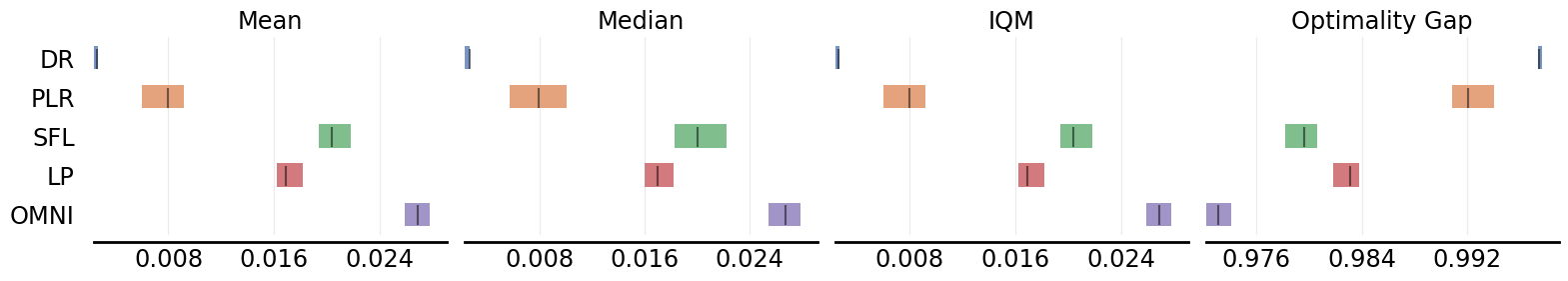}
   
  \caption{95\% Stratified Boostrapped Confidence Intervals for the Mean, Median, Interquartile Mean, and Optimality Gap of Normalized Test Returns of each ACL algorithm on Crafter.}
  \label{fig:crafter_full}
  \vspace{0.5cm}
\end{figure}

\clearpage
\subsection{Neural MMO}
\label{app:nmmo}

Our Neural MMO agents use the exact architecture and hyperparameters provided in the starter kit for the 2023 Neural MMO Competition  \citet{suarez2024neural}. Neural MMO has complex dictionary action and observation spaces, so it uses separate CNN encoders for grid-based observations and fully connected encodings for vector observations, both with ReLU activations, as well as individual fully-connected heads for each action component. These agents are trained with self-play, with a single policy producing batched actions for 128 agents. For the LP, SFL, and OMNI curricula we compute the task reward as $min(max(max_a(R) / 10.0, 0.0), 1.0)$, where $a$ is an agent identifier and $R$ is the mapping of agent identifiers to individual agents' mean episodic return because individual agents get close to but do not exceed 10.0 mean episodic return. We choose to prioritize tasks based on the most successful agent in the environment, but we did not explore other options. It is not clear what the best measure of a successful map is in Neural MMO and so we leave this investigation as future work.

\begin{figure}[h!]
  \centering
  \includegraphics[width=1.0\textwidth]{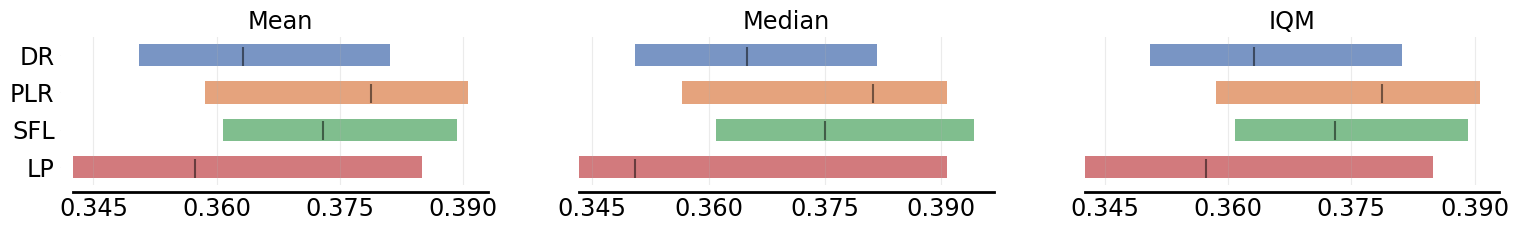}
   
  \caption{95\% Stratified Bootstrapped Confidence Intervals for the Mean, Median, Interquartile Mean, and Optimality Gap of Normalized Test Returns of each ACL algorithm on Neural MMO.}
  \label{fig:nmmo_full}
  \vspace{0.5cm}
\end{figure}

\subsection{NetHack}
\label{app:nethack}
 Our NetHack agents are trained using the open source Moolib code from \citet{hambro2022dungeons}. Moolib implements a version of Asynchronous Proximal Policy Optimization (APPO) \citep{schulman2017proximal, petrenko2020sample}. We use the standard ChaoticDwarfGPT5 baseline from the NetHack Challenge \citep{hambro2022insights}, where the tty character observations are rendered to pixels and passed to the model in addition to text-based information. The model has a CNN with 4 layers and ELU \citep{clevert2015fast} activations for the image observation and linear encoders with ELU  activations for the text components. We use the same training setttings as the NetHackChallenge environment, but we enable seeding to support curriculum learning. Notably, in this environment the episode terminates if the in-game timer does not progress for 150 agent steps, meaning the agent is taking meaningless actions or stuck interacting with menus. For the LP, SFL, and OMNI curricula we compute the task reward as $min(max(R / 1000, 0.0), 1.0)$, where $R$ is the mean episodic return because our agents get close to but do not exceed 1000 mean episodic return.

\begin{figure}[h!]
  \centering
  \includegraphics[width=1.0\textwidth]{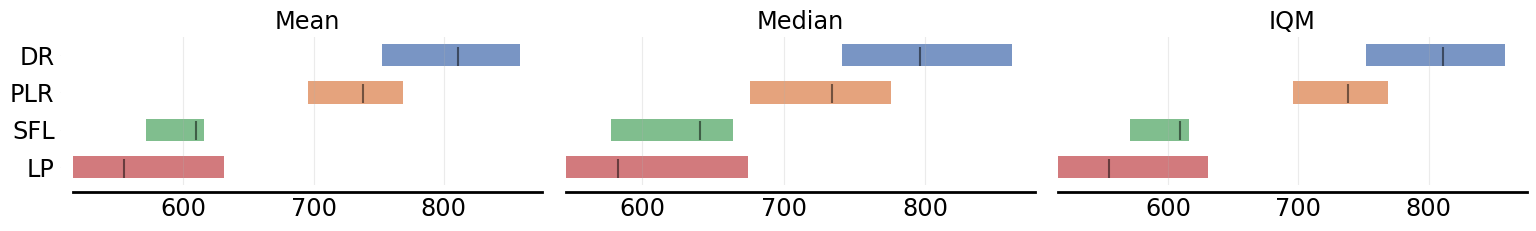}
   
  \caption{95\% Stratified Bootstrapped Confidence Intervals for the Mean, Median, Interquartile Mean, and Optimality Gap of Normalized Test Returns of each ACL algorithm on NetHack.}
  \label{fig:nethack_full}
  \vspace{0.5cm}
\end{figure}

\clearpage
\subsection{Hyperparameters}
\label{app:hyperparameters}
For each baseline in this paper, we use the tuned PPO hyperparameters from the work that introduced the code base. For each environment and curriculum learning method, we perform a separate grid search over the most important hyperparameters (as indicated by their respective papers). For PLR we search for the temperature parameter $\beta$ in {0.1, 0.3, 0.5} and the staleness coefficient $\rho$ in {0.1, 0.3, 0.5}. For LP we search over {0.01, 0.1, 0.2, 0.3, 0.5} for the EMA $alpha$ parameter and over {0.01, 0.1, 0.3} for $p_\theta$. For SFL we try sampling from the full distribution $p * (1 - p)$ and using the SFL method proposed in \citep{rutherfordno} which samples from the top $K$ most learnable tasks with probability $\rho$ and from the full task space with probability $1 - \rho$. We search for $K$ in {10, 25, 50} and $\rho$ in  {0.5, 0.75, 1.0}. Below is the table of all hyperparameters used in this paper.

\begin{table}[h!]
\caption{Learning Hyperparameters.}
\label{table:hyperparams}
\begin{center}
\scalebox{0.87}{
\begin{tabular}{lrrrrr}
\toprule
\textbf{Parameter}             & Procgen  & Crafter  & NetHack  & Neural MMO  & LaserTag \\
\midrule
\emph{PPO}                     &          &          &          &             &          \\
$\gamma$                       & 0.999    & 0.99     & 0.999    & 0.99        & 0.995    \\
$\lambda_{\text{GAE}}$         & 0.95     & 0.95     & 0.95     & 0.95        & 0.95     \\
PPO rollout length             & 256      & 1024     & 32       & 128         & 256      \\
PPO epochs                     & 3        & 4        & 1        & 3           & 5        \\
PPO mini-batches per epoch     & 8        & 16       & 1        & 32          & 4        \\
PPO clip range                 & 0.2      & 0.2      & 0.1      & 0.1         & 0.2      \\
PPO number of workers          & 64       & 32       & 256      & 8           & 32       \\
Adam learning rate             & 5e-4     & 1e-4     & 1e-4     & 1.5e-4      & 1e-4     \\
Adam $\epsilon$                & 1e-5     & 1e-5     & 1e-7     & 1e-6        & 1e-5     \\
PPO max gradient norm          & 0.5      & 0.5      & 0.5      & 0.5         & 0.5      \\
PPO value clipping             & yes      & yes      & yes      & yes         & no       \\
Return normalization           & yes      & no       & yes      & yes         & yes      \\
Value loss coefficient         & 0.5      & 0.5      & 0.5      & 0.5         & 0.5      \\
Entropy coefficient            & 0.01     & 0.01     & 0.001    & 0.01        & 0.0      \\

\addlinespace[10pt]
\emph{PLR}                     &          &          &          &             &          \\
Buffer size, $K$               & 200      & 200      & 200      & 128         & -        \\
Scoring function               & VL1      & VL1      & VL1      & VL1         & -        \\
Prioritization                 & rank     & rank     & rank     & rank        & -        \\
Temperature, $\beta$           & 0.1      & 0.5      & 0.3      & 0.3         & -        \\
Staleness coefficient, $\rho$  & 0.1      & 0.1      & 0.1      & 0.3         & -        \\

\addlinespace[10pt]
\emph{LP}                      &          &          &          &             &          \\
EMA $\alpha$                   & 0.3      & 0.1      & 0.2      & 0.3         & -        \\
Reweighting $p_\theta$         & 0.1      & 0.1      & 0.2      & 0.3         & -        \\
Update period, $T$             & 25       & 25       & 1000     & 50          & -        \\

\addlinespace[10pt]
\emph{SFL}                     &          &          &          &             &          \\
Batch size, $N$                & 200      & 200      & 200      & 128         & -        \\
Update period, $T$             & 25       & 25       & 1000     & 50          & -        \\
Sample method                  & Dist     & Top K    & Dist     & Top K       & -        \\
Buffer size, $K$               & -        & 10       & -        & 25          & -        \\
Sample Ratio, $\rho$           & -        & 1.0      & -        & 0.5         & -        \\

\addlinespace[10pt]
\emph{FSP}                     &          &          &          &             &          \\
Agent checkpoint interval      & -        & -        & -        & -           & 800      \\

\addlinespace[10pt]
\emph{PFSP}                    &          &          &          &             &          \\
$f_{hard}$ entropy coef        & -        & -        & -        & -           & 2        \\
smoothing constant             & -        & -        & -        & -           & 0.01     \\
Win rate episodic memory       & -        & -        & -        & -           & 128      \\

\bottomrule 
\end{tabular}
}
\end{center}
\end{table}

\clearpage
\section{Additional Experiments}

\subsection{RLLib with Syllabus}
\label{app:cartpole-rllib}
Cart Pole is a toy environment mainly used to debug RL implementations \citep{6313077}. It initializes a cart to a random starting point along a 2D track and tasks the agent with balancing a pole for as long as possible. It is not a multitask environment, so we use Syllabus's task wrapper to make the cart's initialization range a configurable option. This experiment demonstrates how Syllabus's Task Interface can add multitask functionality to singleton environments, and how Syllabus integrates easily with RLLib's Ray-based multiprocessing \citep{moritz2018ray, liang2018rllib}.

\begin{figure}[h!]
  \centering
    \centering
    \includegraphics[width=0.8\textwidth]{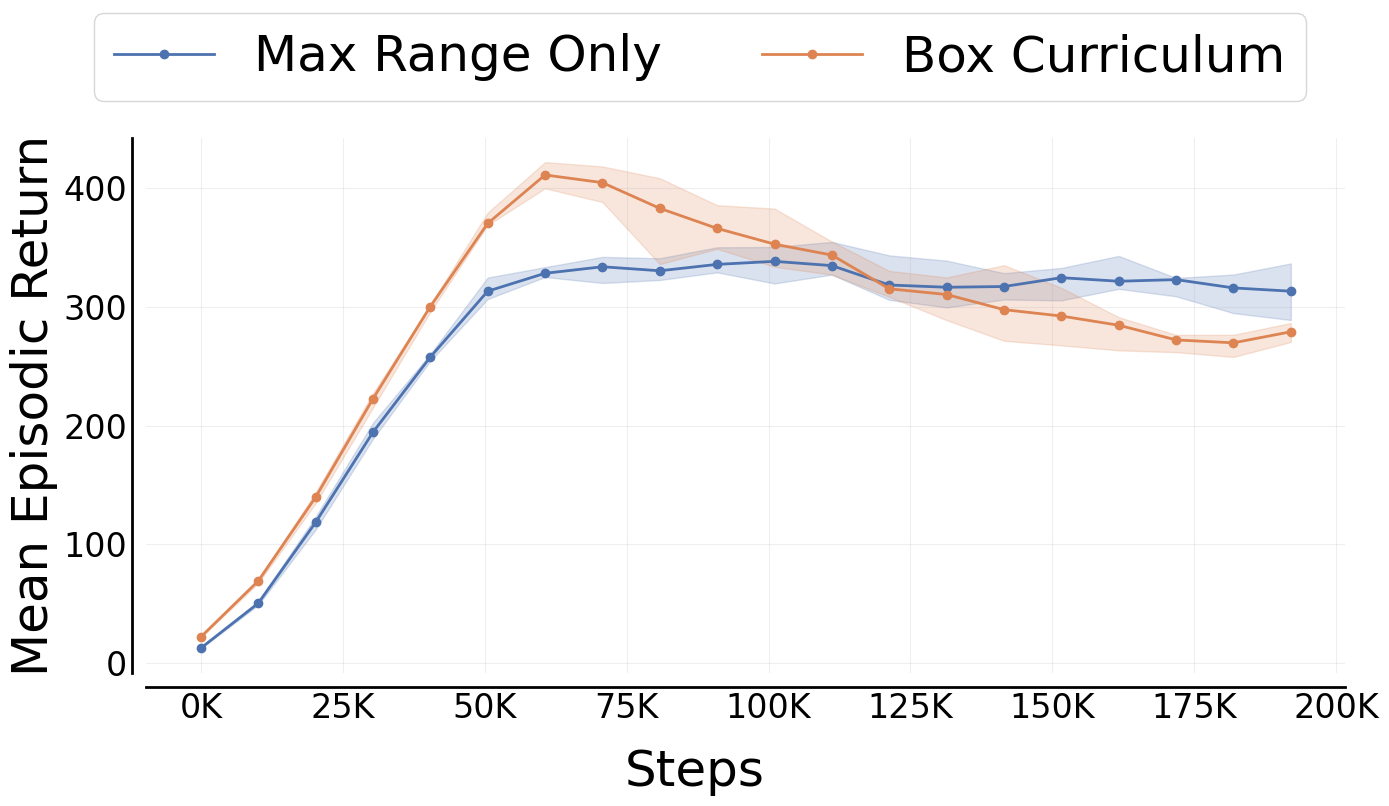}
    \subcaption{Training Cart Pole agents in RLLib, with curricula over the range of possible initial cart positions. We compare a simple curriculum of expanding the range throughout training vs. training with the fixed maximum initialization range.}
    \label{subfig:rllib_cartpole}
\end{figure}

Our Cart Pole experiments use a simple curriculum that increases the initialization range of the cart over the course of training. This causes the cart to begin in more precarious positions. We compare this to an agent trained only with the maximum initialization range. The curriculum learning agent initially learns a strong policy, but converges to a weaker policy than the agent trained solely on the maximum range, as we see in \autoref{subfig:rllib_cartpole}. Since the single-task agent easily converges to a strong policy, there is little reason to use curriculum learning.

\clearpage
\subsection{Self-Play on LaserTag}
\label{app:lasertag}

\begin{figure}[h!]
  \centering
  \begin{subfigure}[t]{0.48\textwidth}
      \centering
      \includegraphics[width=1.0\textwidth]{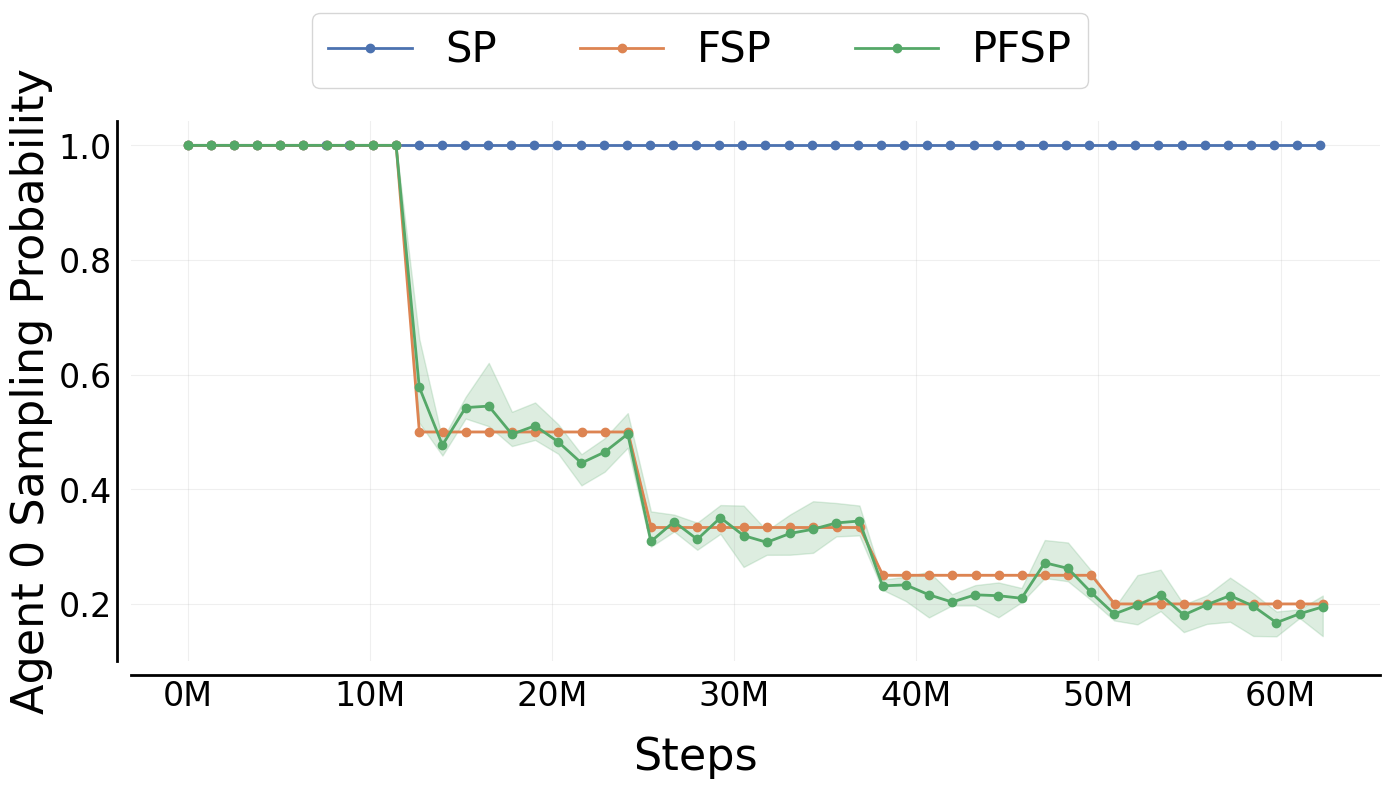}
      \caption{}
      \label{fig:lasertag_prob}
  \end{subfigure}
  \begin{subfigure}[t]{0.48\textwidth}
      \centering
      \includegraphics[width=1.0\textwidth]{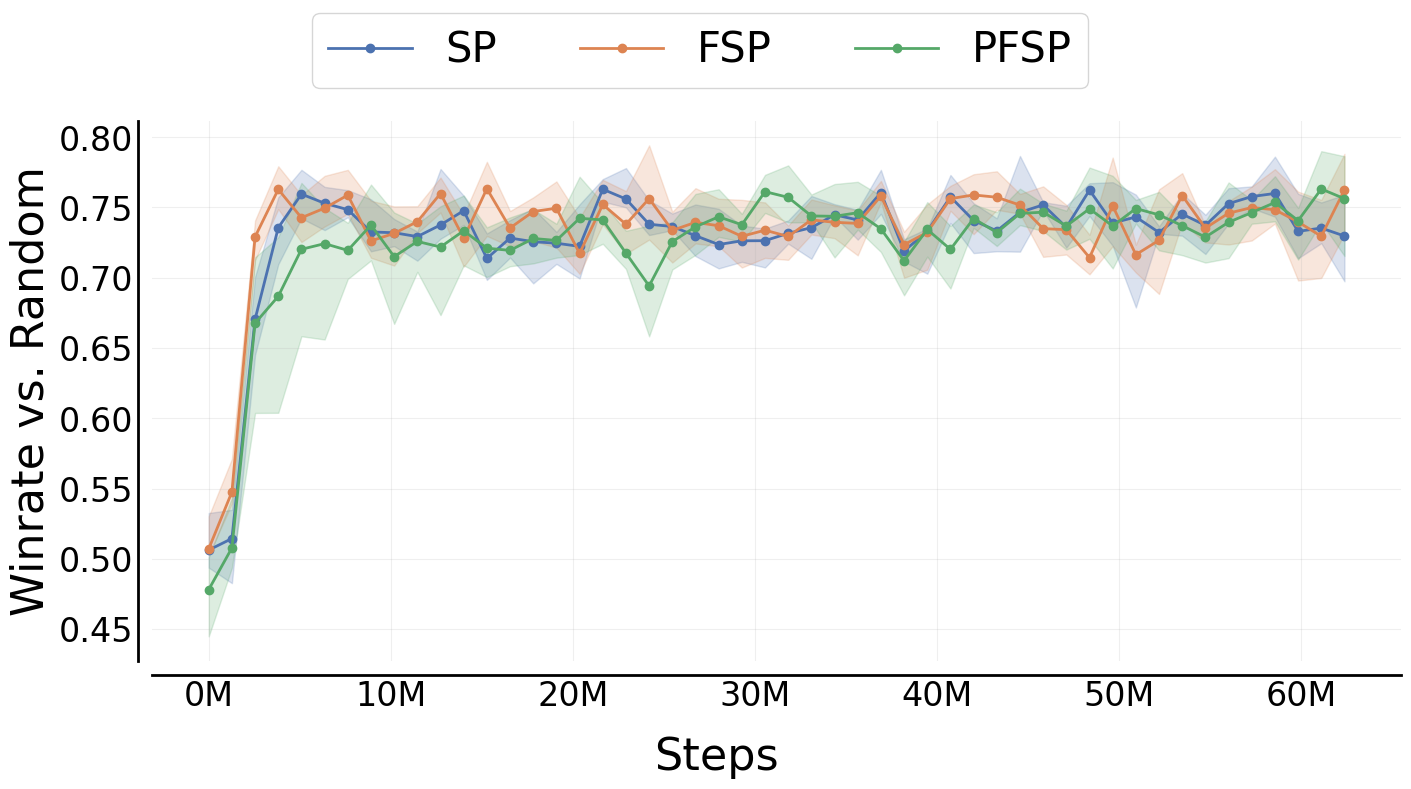}
      \caption{}
      \label{fig:lasertag_win}
  \end{subfigure}
  \begin{subfigure}[t]{1.0\textwidth}
    \includegraphics[width=1.0\textwidth]{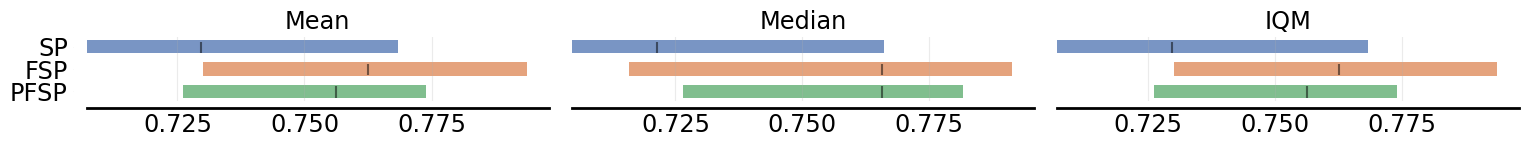}
  \end{subfigure}
  \caption{\textbf{(a)} Probability of sampling the first agent for each algorithm. Self-play only has one agent, so it always samples the current policy as it's opponent. \textbf{(b)} The winrate of each method against a fixed random policy. \textbf{(c)} Stratified bootstrapped confidence intervals of the winrates.}
  \label{fig:lasertag}
\end{figure}

We test our self-play algorithms (Self-Play, Fictitious Self-Play, and Prioritized Fictitious Self-Play) on the LaserTag environment introduced by \citet{lanctot2017unified} and implemented by \citet{samvelyan2022maestro} in Griddly \citep{bamford2021griddly}. We use the same model architecture and hyperparameters as \citet{samvelyan2022maestro} and train our agents for 4000 updates or 65,536,000 environment steps, and add a copy of the current policy to the opponent buffer after every 800 updates. SP always plays against the current agent (agent 0), FSP uniformly samples from all past agents, and PFSP samples from all past agents based on the current agent's winrate against them. We see in \autoref{fig:lasertag_prob} that our methods properly update their sampling distributions after each agent is added to the opponent buffer. However, we do not see any difference between the methods against a fixed random agent in \autoref{fig:lasertag_win}. We may need to train for longer to see any benefits to sampling historic agents.

\clearpage
\subsection{Phasic Policy Gradients}
\label{app:ppg}

\begin{figure}[h!]
  \centering
  \begin{subfigure}[t]{0.8\textwidth}
      \centering
      \includegraphics[width=1.0\textwidth]{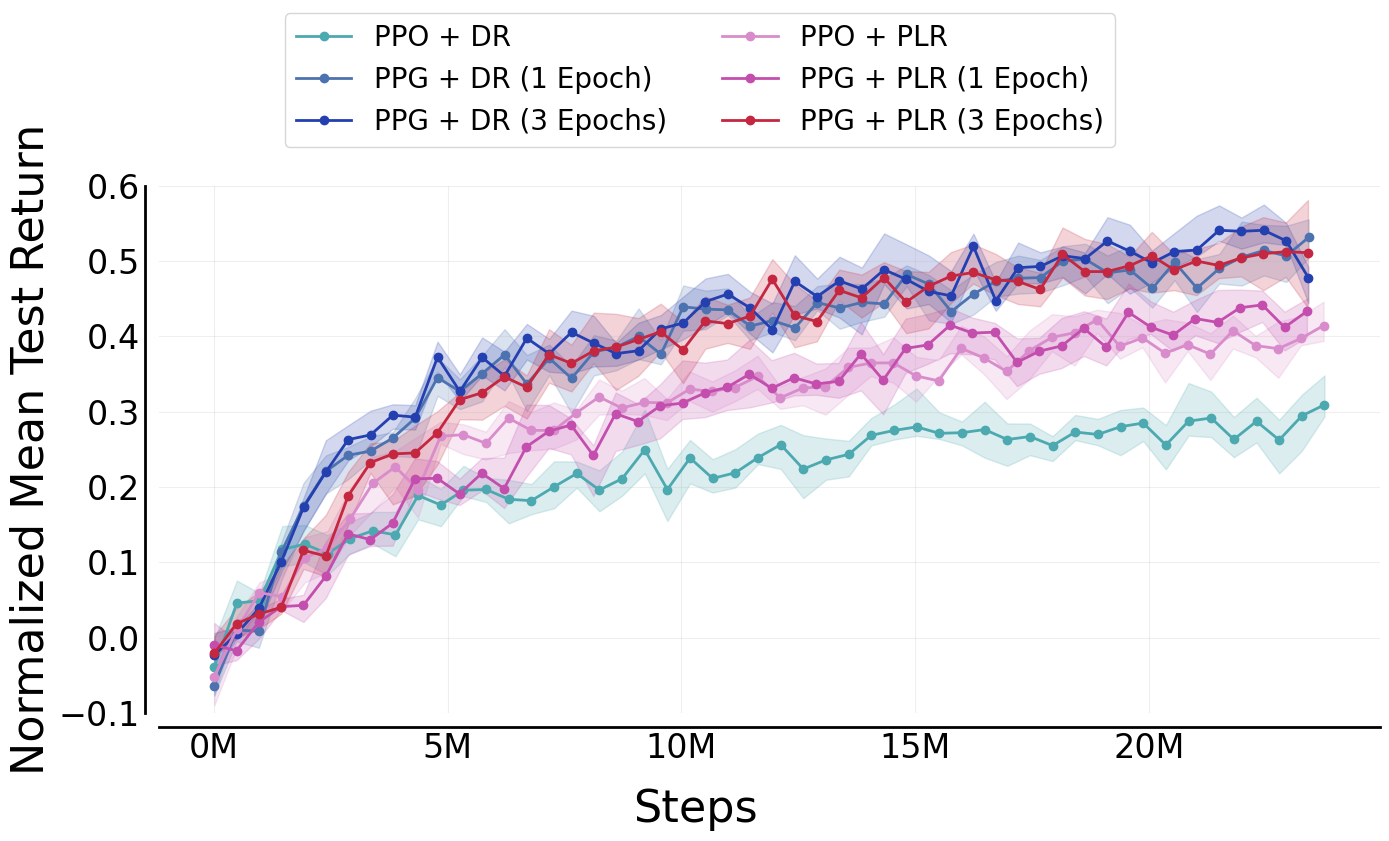}
  \end{subfigure}
  \begin{subfigure}[t]{1.0\textwidth}
    \includegraphics[width=1.0\textwidth]{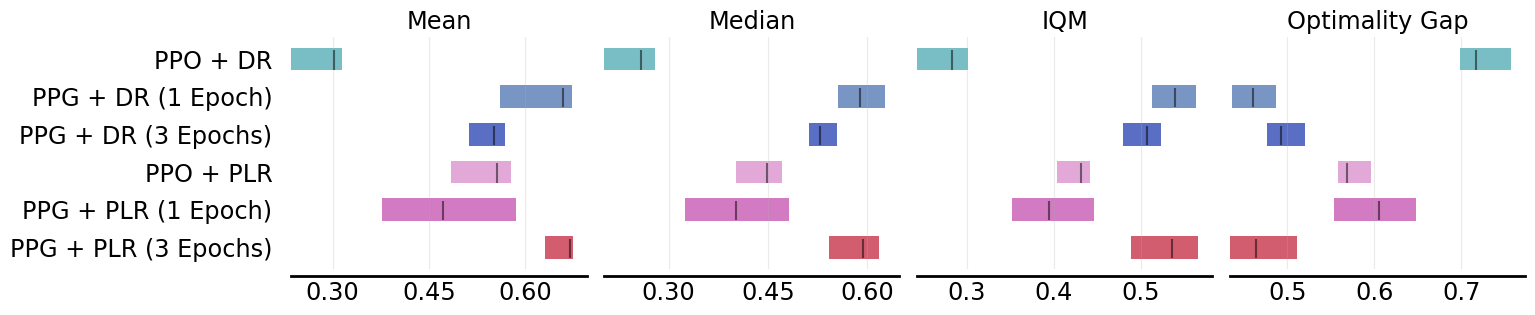}
  \end{subfigure}
  \caption{Normalized Test Returns of PPO and PPG with 1 or 3 value epochs when trained with DR or PLR on 10 Procgen environments with 5 seeds each.}
  \label{fig:ppg}
\end{figure}

Phasic Policy Gradients (PPG) \citep{cobbe2021phasic} is an extension of PPO that trains a separate value and policy network, while distilling features from the value network into the actor. They do this by adding an auxiliary value head to the policy, and periodically using behavior cloning from the value network into the auxiliary value head, allowing the actor to learn features used critic. \citet{cobbe2021phasic} showed that this approach outperforms PPO on Procgen. We use Syllabus study whether Prioritized Level Replay can provide an additional level of improvement over PPG as it does for PPO. PPG updates the policy and value with a separate number of epochs, but by default uses 1 value epoch and 1 policy epoch. We see in \autoref{fig:ppg} that applying PLR to PPG with the default hyperparameters performs worse than DR, despite its close similarity to PPO. We hypothesize that this is may be due to the lack of value updates. If the value predictions are less accurate, then PLR's score will also be inaccurate. We further investigate increasing the number of value epochs, and find that by increasing the number of value epochs to 3, the same as PPO, PLR matches but does not exceed DR.

\clearpage
\subsection{Stale Value Predictions}
\begin{figure}[h!]
  \centering
  \begin{subfigure}[t]{0.8\textwidth}
      \centering
      \includegraphics[width=1.0\textwidth]{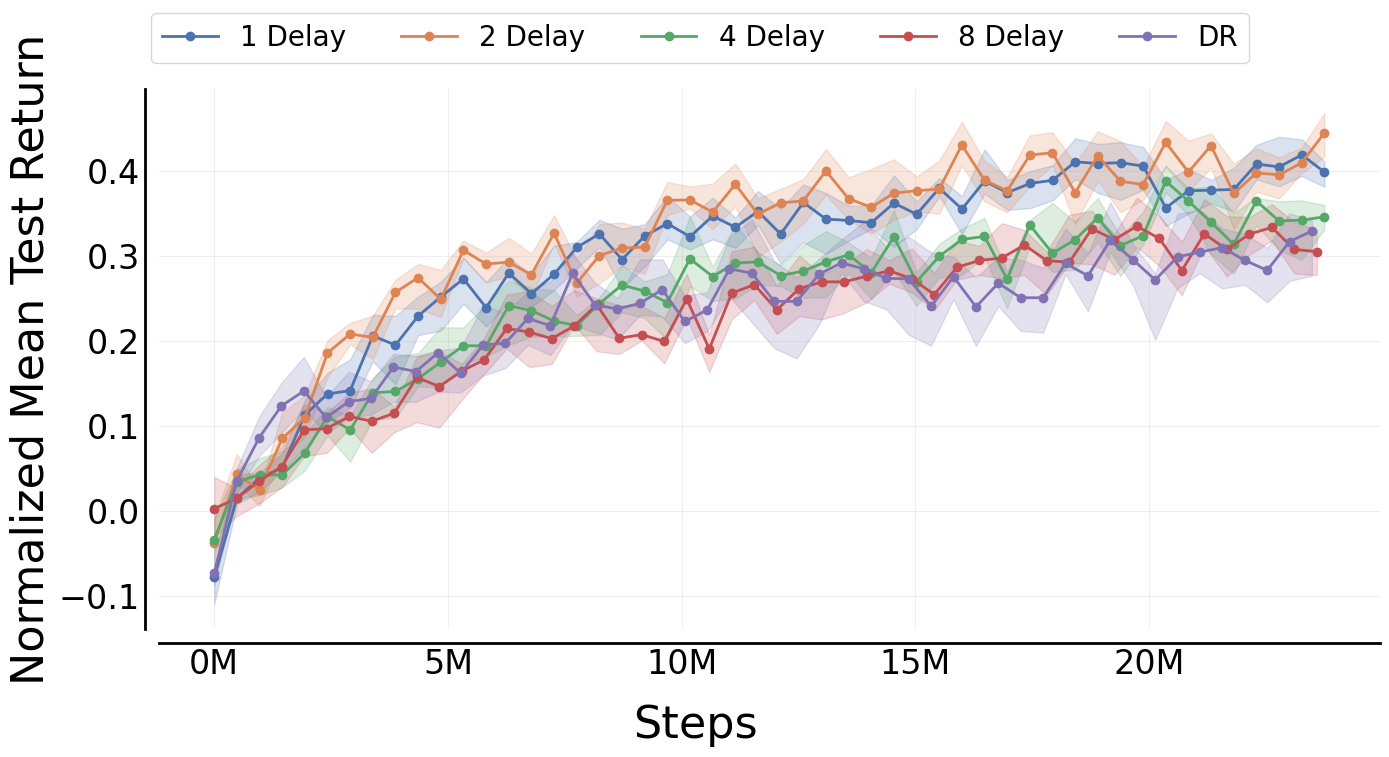}
  \end{subfigure}
  \begin{subfigure}[t]{1.0\textwidth}
    \includegraphics[width=1.0\textwidth]{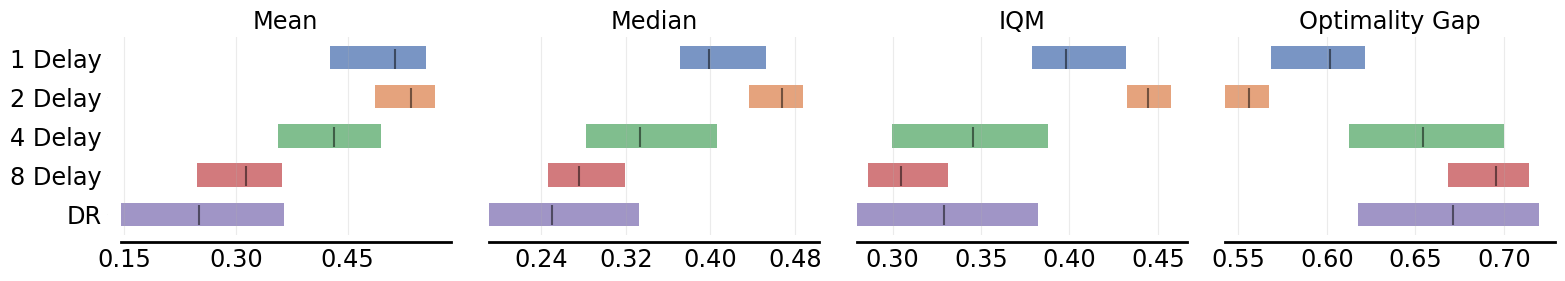}
  \end{subfigure}
  \caption{Training Procgen agents with PLR using stale value predictions. 1 buffer is equivalent to 1 episode of delay per environment.}
  \label{fig:ppg-buffer}
\end{figure}

We continue the investigation from \autoref{app:ppg} and further explore the importance of value prediction quality by training on delayed tasks with PLR. This is equivalent to sampling from a distribution calculated from stale value predictions. We can use Syllabus's asynchronous sampling code to artificially increase the delay. We find in \autoref{fig:ppg-buffer} that as the delay increases, the performance of PLR drops until it is nearly equivalent to domain randomization.

\clearpage
\subsection{Sampling for Learnability with OMNI}
\label{app:omni-learnability}
\begin{figure}[h!]
  \centering
  \begin{subfigure}[t]{0.8\textwidth}
      \centering
      \includegraphics[width=1.0\textwidth]{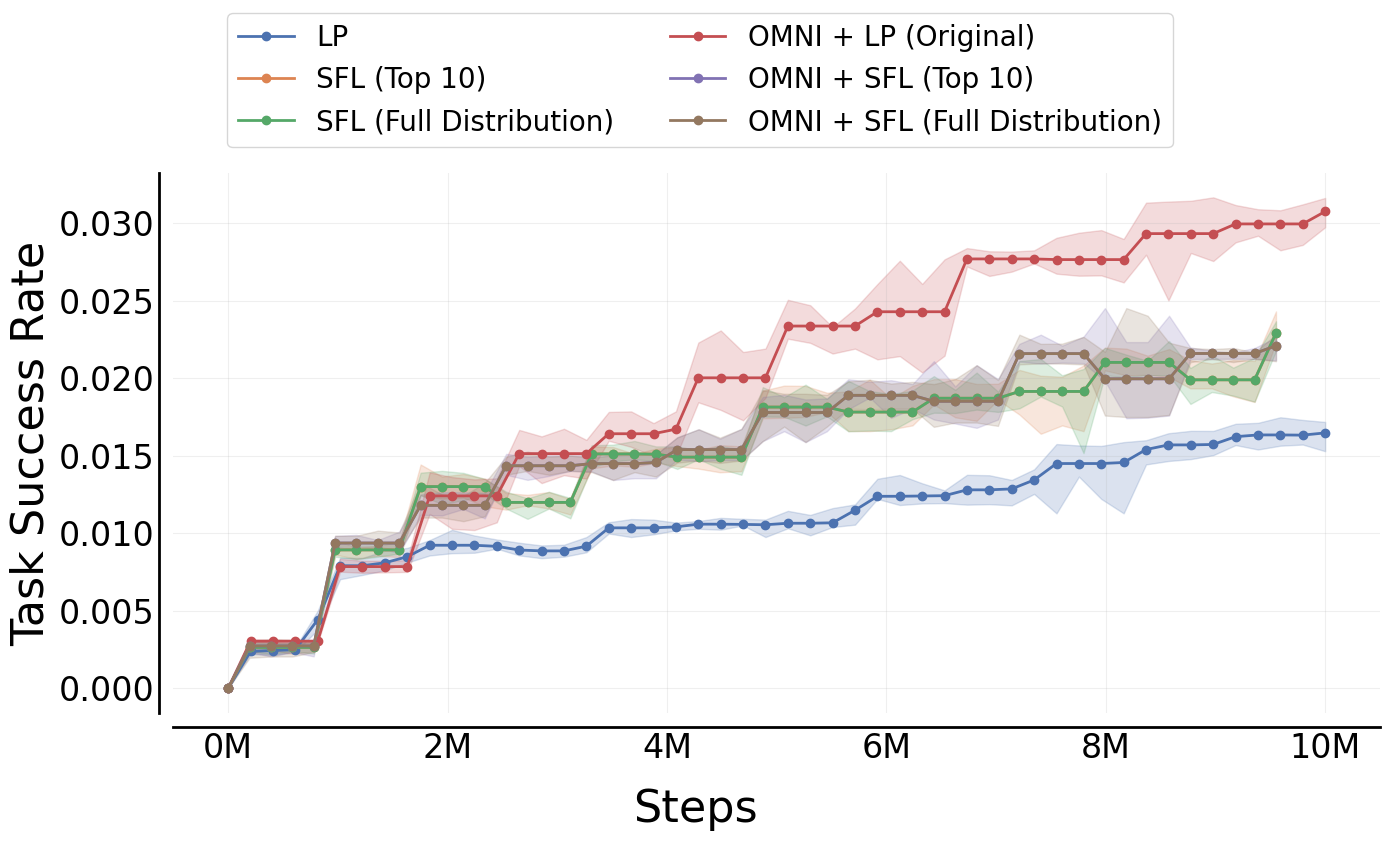}
  \end{subfigure}
  \begin{subfigure}[t]{1.0\textwidth}
    \includegraphics[width=1.0\textwidth]{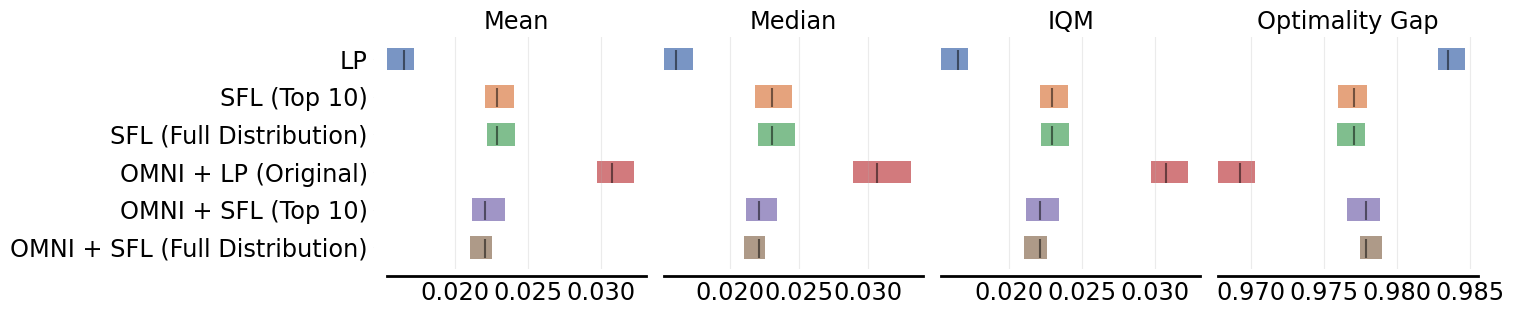}
  \end{subfigure}
  \caption{Mean task success rates for the Full Distribution and Top K implementations of SFL with OMNI's interestingess filter. LP and OMNI are shown for reference.}
  \label{fig:omni-learnability}
\end{figure}

Sampling for Learnability and Learning Progress are very similar methods in terms of implementation. Both periodically evaluate the agent over the entire task space to generate task success rates, then generate a sampling distribution from those rates. OMNI adds an additional component to the LP curriculum by using an LLM to filter non-interesting tasks out of the distribution. In \autoref{fig:crafter} we found that SFL outperforms Learnability on Crafter, so we chose to investigate whether OMNI's filtering step could also improve the performance of SFL. Unfortunately we find that neither the full distribution nor top K implementations of SFL benefit from OMNI's filtering, as seen in \autoref{fig:omni-learnability}.

\clearpage
\subsection{Neural MMO Task-Based Curriculum}
\label{app:nmmo-sequential}

We rely on events and achievements that are not zero-sum to determine how proficient each agent is in the environment, but these metrics are not completely independent from the quality of opponents. For example, the frequency of dying to starvation may decrease because agents are becoming better at scavenging for food, or because they are becoming more proficient at fighting, increasing their chance of dying to combat rather than starvation. 

The sequential curriculum consists of 5 stages, each of which uses domain randomization over a subset of the task space. The curriculum progresses to the next stage whenever the agent achieves a mean episodic return of 0.75 averaged over the past 1000 episodes. Each agent successfully made it to the final curriculum stage by the end of training.

For each individual task, we assign a threshold. The agent is gets a reward of 1.0 for completing the task, which is distributed as the agent makes progress on the task. For example, if we task the agent with surviving for 100 timesteps, after 50 timesteps it will have a cumulative reward of 0.5. After surviving for 150 timesteps, the agent will have a cumulative reward of 1.0 because we stop assigning reward after the threshodl is reached. The tasks and thresholds for each stage are listed in \autoref{tab:nmmo_curriculum}

\begin{table}[h!]
    \centering
    \caption{Neural MMO Sequential Curriculum Task Thresholds}
    \begin{tabular}{@{}lccccc@{}}
        \toprule
        Task & Stage 1 & Stage 2 & Stage 3 & Stage 4 &  Stage 5  \\
        \midrule
        Survive (steps)         & 50 & 150 & 250 & 350 & 500 \\
        Eat Food (count)        & 5 & 15 & 25 & 35 & 50 \\
        Drink Water (count)     & 5 & 15 & 25 & 35 & 50 \\
        Harvest Item (count)    & 3 & 9 & 15 & 21 & 30 \\
        Go Far (distance)       & 3 & 9 & 15 & 21 & 30 \\
        Level Up (count)        & 2 & 6 & 10 & 14 & 20 \\
        Equip Item (count)      & 1 & 3 & 5 & 7 & 10 \\
        Consume Item (count)    & 1 & 3 & 5 & 7 & 10 \\
        Buy Item (count)        & 1 & 3 & 5 & 7 & 10 \\
        Player Kill (count)     & 1 & 3 & 5 & 7 & 10 \\

        \bottomrule
    \end{tabular} 
    \vspace{0.5cm}
    \label{tab:nmmo_curriculum}
\end{table}

\begin{figure}[h!]
  \centering
  \includegraphics[width=1.0\textwidth]{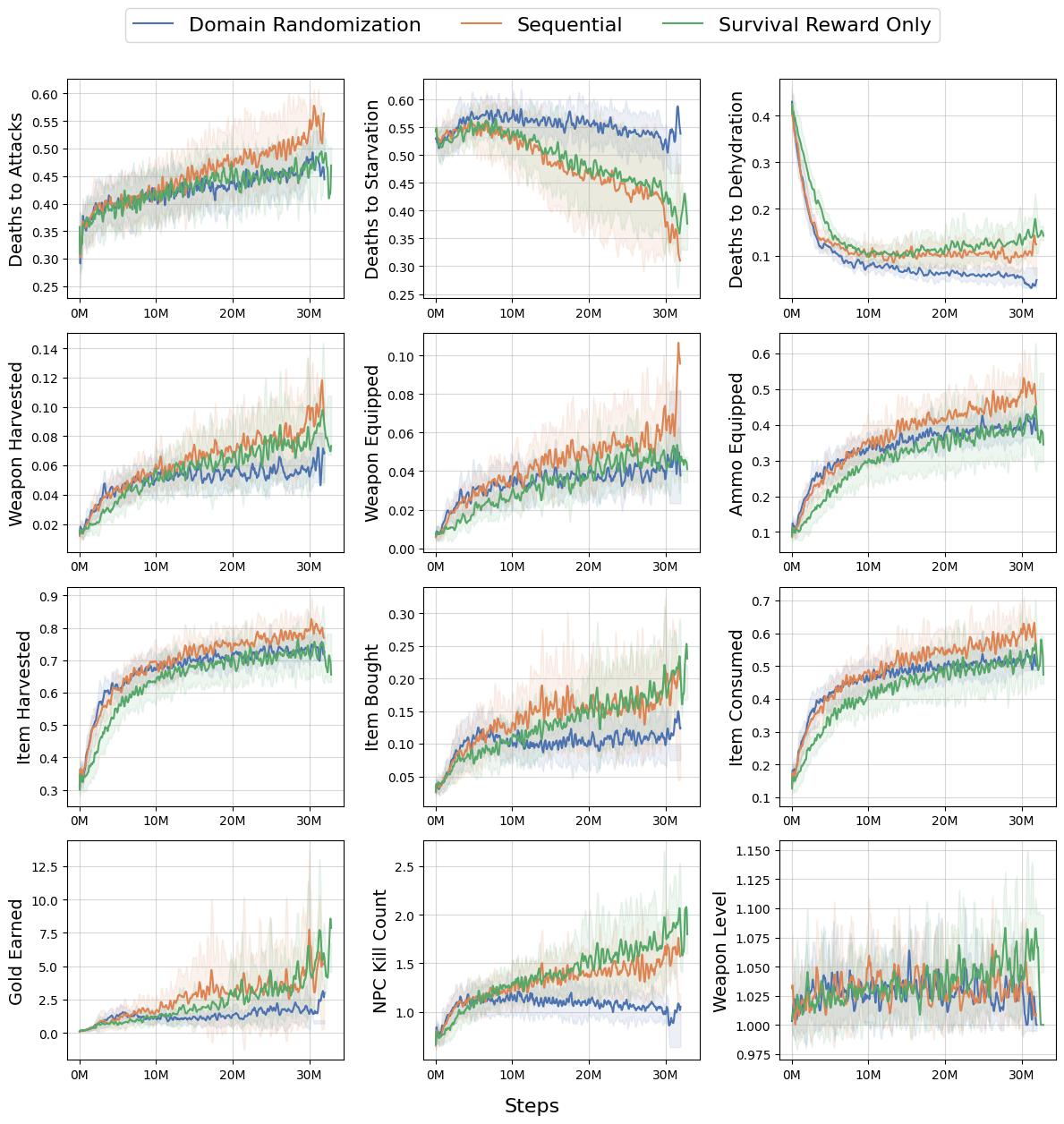}
   
  \caption{Various events and achievements in Neural MMO for Domain randomization and a manually designed sequential curriculum over the course of training.}
  \label{fig:nmmo-sequential}
\end{figure}

We see in \autoref{fig:nmmo-sequential} that the sequential curriculum seems to create more aggressive before as these agents harvest and equip more weapons and ammo, and die to attacks more frequently. On the other hand, agents trained with domain randomization utilize the weapon and item system at similar rates to the other agents, but fail to learn some game systems like earning gold and killing NPCs.

\clearpage
\section{Code Examples}
\subsection{RLLib}

\begin{figure}[!ht]
\caption{Adding curriculum learning with Syllabus to RLLib training code with just a few lines of code.}
\label{app:rllib}
\begin{minted}[frame=single, linenos]{diff}
    import gym
    from ray.tune.registry import register_env
    from ray import tune
    from gym.spaces import Box
    from .task_wrappers import CartPoleTaskWrapper

+   from syllabus.core import RaySyncWrapper, make_ray_curriculum
+   from syllabus.curricula import ExpandingBox
+   from syllabus.task_space import BoxTaskSpace

    if __name__ == "__main__":
+       # Define a task space
+       task_space = BoxTaskSpace(Box(-0.3, 0.3, shape=(2,)))

        def env_creator(config):
            env = gym.make("CartPole-v1")
+           # Wrap the environment to change tasks on reset()
+           env = CartPoleTaskWrapper(env)
+           # Add environment sync wrapper
+           env = RaySyncWrapper(env, task_space)
            return env

        register_env("task_cartpole", env_creator)

+       # Create the curriculum
+       curriculum = ExpandingBox(task_space)
+       # Add the curriculum sync wrapper
+       curriculum = make_ray_curriculum(curriculum)

        config = {
            "env": "task_cartpole",
            "num_gpus": 1,
            "num_workers": 8,
            "framework": "torch",
        }

        tuner = tune.Tuner("APEX", param_space=config)
        results = tuner.fit()

\end{minted}
\end{figure}

\clearpage
\subsection{Stable Baselines 3}
\label{app:sb3}
\begin{figure}[ht!]
\begin{minted}[frame=single, linenos]{diff}
    import gym
    import procgen  # noqa: F401
    from stable_baselines3 import PPO
+   from syllabus.core import make_multiprocessing_curriculum
+   from syllabus.curricula import DomainRandomization
+   from syllabus.examples.task_wrappers import ProcgenWrapper
+   from syllabus.task_space import TaskSpace

    if __name__ = "__main__":
        def make_env(curriculum, task_space):
            def thunk():
                env = gym.make("procgen-bigfish-v0")
+               # Wrap the environment to change tasks on reset()
+               env = ProcgenWrapper(env)
+               # Add environment sync wrapper
+               env = GymnasiumSyncWrapper(
+                   env,
+                   task_space,
+                   curriculum.components,
+               )
                return env
            return thunk

+       # Define a task space
+       task_space = DiscreteTaskSpace(200)
+       # Create the curriculum
+       curriculum = DomainRandomization(task_space)
+       curriculum = make_multiprocessing_curriculum(curriculum)

        venv = DummyVecEnv(
            [
                make_env(curriculum, task_space)
                for i in range(64)
            ]
        )

        model = PPO("CnnPolicy", venv)
        model.learn(25000000)

\end{minted}
\end{figure}

\clearpage
\pagebreak

\section{Documentation}
\label{app:documentation}
\begin{figure}[h!]
  \centering
  \includegraphics[width=1.0\textwidth]{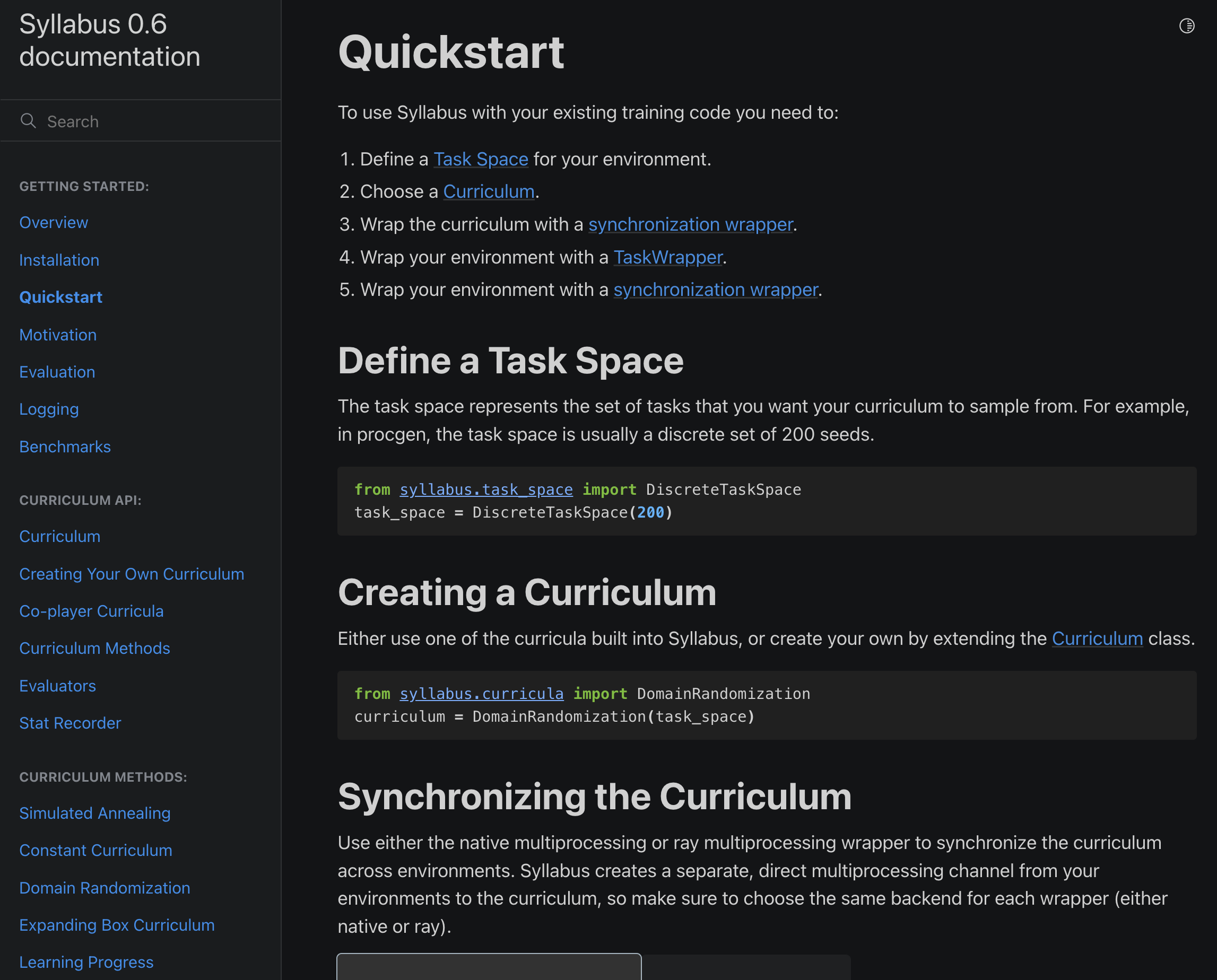}
   
  \caption{Quickstart page of Syllabus's documentation website.}
  \label{fig:documentation}
\end{figure}

Syllabus is documented both in code and with a dedicated documentation website. This includes details on each curriculum algorithm and warnings about common pitfalls that users might run into when configuring them. A sample of the documentation website is shown in \autoref{fig:documentation}.

\end{document}